\documentclass{article}

\usepackage{microtype}
\usepackage{graphicx}
\usepackage{subcaption}
\usepackage{booktabs} 

\PassOptionsToPackage{hyphens}{url}  
\usepackage{hyperref}
\usepackage{xurl}  

\usepackage[preprint]{icml2026}

\usepackage[utf8]{inputenc}
\usepackage[T1]{fontenc}
\usepackage{mathtools}
\usepackage{amsmath}
\usepackage{amssymb}
\usepackage{amsfonts}
\usepackage{amsthm}
\usepackage{nicefrac}
\usepackage{enumitem}
\usepackage{xcolor}
\usepackage[most]{tcolorbox}
\usepackage{xspace}
\usepackage{float}   
\usepackage{caption}
\usepackage{tikz}

\usepackage[capitalize,noabbrev]{cleveref}

\usepackage{algorithm}
\usepackage{algorithmic}

\theoremstyle{plain}

\theoremstyle{definition}

\theoremstyle{remark}

\newcommand{\topk}{\text{top-}k}

\newcommand{\codebutton}[1]{%
  \tikz[baseline=(X.base)]{
    \node[
      fill=gray!18,
      draw=gray!35,
      rounded corners=1.3pt,
      inner xsep=2.2pt,
      inner ysep=0.2pt,
      text height=1.5ex,
      text depth=0.25ex,
      font=\ttfamily
    ] (X) {#1};
  }%
  \xspace%
}

\newcommand{\Drop}{\codebutton{Drop}}
\newcommand{\Add}{\codebutton{Add}}
\newcommand{\Flip}{\codebutton{Flip}}

\newcommand{\PertType}{\textit{perturbation type}}
\newcommand{\TopkMem}{\textit{top-$k$ membership}}
\newcommand{\KTau}{\textit{Kendall's $\tau$}}
\newcommand{\CIUncert}{\textit{confidence-interval-based uncertainty}}

\icmltitlerunning{A Unified Perturbation Framework for Leaderboard Stability and Manipulation}

\begin{document}

\twocolumn[
  \icmltitle{A Unified Perturbation Framework for\\
    Analyzing Leaderboard Stability and Manipulation}

  \icmlsetsymbol{equal}{*}

\begin{icmlauthorlist}
  \icmlauthor{Hosna Oyarhoseini}{uw}
  \icmlauthor{Jimmy Lin}{uw}
  \icmlauthor{Amir-Hossein Karimi}{uw}
\end{icmlauthorlist}

\icmlaffiliation{uw}{University of Waterloo, Waterloo, Ontario, Canada}

\icmlcorrespondingauthor{Hosna Oyarhoseini}{hoyarhos@uwaterloo.ca}
  \icmlkeywords{Machine Learning, Bradley-Terry, Leaderboards, Influence Functions, Robustness}

  \vskip 0.3in
]

\printAffiliationsAndNotice{}  

\begin{abstract}
Evaluation leaderboards such as LMArena play a central role in benchmarking large language models by aggregating pairwise human preferences into model rankings, yet the robustness of these rankings remains poorly understood.
We present a unified perturbation framework for analyzing Bradley--Terry leaderboards under structured data modifications using influence-based approximations.
Our framework studies three match-level perturbations---\Drop, \Add, and \Flip---together with player removal, and evaluates their effects on \TopkMem, global ranking consistency via \KTau, and \CIUncert.
Across Chatbot Arena and six additional pairwise-comparison datasets, we show that modern leaderboards are non-robust across all three objectives: sub-1\% targeted perturbations can change the top-ranked model, degrade \KTau, and alter confidence intervals.
Beyond robustness auditing, we show that the same influence scores enable efficient targeted perturbations, promoting or demoting specific models and reducing target-model uncertainty with fewer actions than previous manipulation and active-sampling baselines.
By summarizing these effects with normalized dataset-level robustness scores, our framework provides a practical and helpful tool for auditing leaderboard stability and motivating more robust evaluation protocols.
\end{abstract}
\section{Introduction}

The rapid proliferation of Large Language Models (LLMs) has necessitated 
the development of scalable, human-centric evaluation {frameworks}~\citep{frick2025prompttoleaderboard, miroyan_search_2025, zheng2023judging, guo2023evaluatinglargelanguagemodels}. 
Because the quality of open-ended generation is subjective and hard to capture with a single absolute metric, modern platforms have increasingly adopted pairwise preference comparisons, where judges choose between two model outputs. In systems such as LMArena/Chatbot Arena, these preferences are primarily collected from human voters \citep{chiang2024chatbot}, though LLM-based judges are also increasingly used in related evaluation settings~\citep{zheng2023judging}. 
This paradigm builds on a classical approach for ranking items from direct
comparison outcomes~\citep{bradley_rank_1952, negahban_rank_2015, wauthier_efficient_2013, JMLR:v18:16-206}, 
but has recently become a central pillar of LLM benchmarking~\citep{zheng2023judging, chiang2024chatbot}. 
Most notably, Chatbot Arena~\citep{chiang_chatbot_2024} uses Bradley--Terry scores
derived from crowdsourced votes to produce model rankings.
Beyond leaderboard evaluation, Bradley--Terry models are also underpinned to reward
model training for RLHF~\citep{ouyang_training_2022, bai_training_2022,
lee_rlaif_2023, touvron_llama_2023, xu_uncertainty_2024,
sun_rethinking_2025} and have been used to route queries to suitable LLMs or
inference-time scaling strategies~\citep{damani_learning_2025}. As these
leaderboards increasingly shape model adoption and industry recognition
~\citep{metz2025deepseekchatbotarena, kruppa2024berkeleyarena,
singh_leaderboard_2025}, they are often treated as stable estimates of true
model skill.

Recent work challenges this assumption~\citep{huang_dropping_2025, 
min_improving_2025, huang_exploring_2025, singh_leaderboard_2025, 
wu_diagnostic_2022}. 
\citet{huang_dropping_2025} show that removing only a small number of pairwise 
comparisons can alter top rankings, while other studies identify vulnerabilities 
from injected votes~\citep{min_improving_2025}, gamed LLM judges~\citep{zheng_cheating_2025, raina_llm_judge_2024}, 
apathetic or arbitrary annotators~\citep{zhao_challenges_2025}, and data 
leakage or selective reporting~\citep{singh_leaderboard_2025}. 
However, existing analyses typically focus on a single \PertType{} or 
ranking objective. 
This leaves open a broader question: \emph{how do different small, structured changes to 
the comparison dataset propagate through different leaderboard outcomes?} 
We address this question by building on \citep{huang_dropping_2025}'s idea with a unified influence-based perturbation framework 
for Bradley--Terry leaderboards. Our framework treats dataset modifications as 
structured interventions and propagates their effects through the ranking 
estimator to downstream leaderboard conclusions. This allows us to study both 
robustness failures and targeted interventions within the same formalism, 
including match-level perturbations and player-level removal motivated by model 
deprecation or exclusion.

Our main contributions are:
\begin{itemize}[leftmargin=*,topsep=2pt,itemsep=1pt]
    \item We introduce a unified influence-based framework for auditing 
Bradley--Terry leaderboards under three match-level perturbations---\Drop{}, 
\Add{}, and \Flip{}---instantiated for three ranking objectives: \TopkMem, 
\KTau{} for global consistency, and \CIUncert{} for the reliability of 
estimated skills.
    
    \item Across seven pairwise-comparison datasets, we show that leaderboard 
    non-robustness is systematic across datasets, \PertType{}, and ranking 
    criteria: fewer than 1\% of comparisons substantially affect \TopkMem, 
    \KTau, and CI-based stability. We also aggregate influence-guided failures into normalized dataset-level robustness
    scores, giving a compact audit profile across Top-$k$, CI-aware, and
    global-ranking views.
    
    \item The same framework supports targeted interventions, promoting or 
    demoting specific models with fewer actions than prior 
    vote-manipulation~\citep{min_improving_2025} and identifying matchups that 
    reduce uncertainty more effectively than Chatbot Arena-style active 
    sampling~\citep{chiang_chatbot_2024}. 
    
    \item A player-removal analysis shows that removing
    influential players can induce broad reordering, highlighting model
    deprecation as a source of the leaderboard illusion~\citep{singh_leaderboard_2025}.
    
\end{itemize}

Overall, this work provides a unified perspective on leaderboard non-robustness by bridging robustness analysis and adversarial manipulation, revealing fundamental limitations in current leaderboard designs and highlighting the need for more reliable and trustworthy benchmarking methodologies.

\section{Preliminaries}
\label{sec:preliminaries}

\subsection{Pairwise comparison and Bradley--Terry framework}

Modern LLM leaderboards rank models from pairwise preference data rather than absolute scores. In this setting, each \emph{player} denotes an entity being ranked; for LLM leaderboards, a player corresponds to an LLM or model. A \emph{match} or \emph{vote} denotes one pairwise comparison between two players, typically obtained when a human or judge compares two model responses to the same prompt and selects the preferred output. These comparisons are 
commonly modeled using the Bradley--Terry (BT) framework~\citep{bradley_rank_1952}, 
which estimates latent skill scores from pairwise outcomes.

We adopt the notation of \cite{chiang2024chatbot}, defining a set of $M$ models $\mathcal{M} = \{m_1, \dots, m_M\}$ and a dataset $D = \{z_n\}_{n=1}^N$. Each observation $z_n = (x_n, y_n)$ compares two players $(i_n,j_n)$, where $x_n=e_{i_n}-e_{j_n}\in\mathbb{R}^M$ encodes the matchup and $y_n \in \{0, 1\}$ indicates whether $i_n$ beats $j_n$. For a comprehensive breakdown of the data format and our specific protocol for handling ties, please refer to Appendix \ref{app:bt-io-format}.

The BT model assigns each model $m_i$ a scalar latent skill coefficient $\theta_i \in \mathbb{R}$. 
For comparison \(n\), let \((i,j)\) denote the ordered pair of models being
compared, and let \(y_n=1\) indicate that model \(i\) is preferred to model \(j\).
The Bradley--Terry probability is
\begin{equation}
\begin{aligned}
    p_n &:= P(y_n = 1 \mid i,j;\theta) \\
    &= \sigma(x_n^\top\theta) = \sigma(\theta_i - \theta_j) \\
    &= \frac{1}{1 + e^{-(\theta_i - \theta_j)}} .
\end{aligned}
    \label{eq:BT}
\end{equation}

To ensure identifiability against constant shifts, we fix $\theta_1 = 0$. The parameter vector $\theta$ is estimated by minimizing the empirical binary cross-entropy loss:
\begin{equation}
\begin{aligned}
\hat{\theta} &= \arg\min_{\theta} \sum_{n=1}^N \ell(z_n;\theta) \\
&= \arg\min_{\theta} \sum_{n=1}^N \big[-y_n \log p_n \\
&\qquad\qquad - (1-y_n)\log(1-p_n)\big].
\end{aligned}
\end{equation}

\subsection{Influence functions and propagation to objectives}
Previous work has used influence scores to identify high-impact deletions under
a fixed budget~\citep{broderick_automatic_2023}. Building on this idea, we study
how perturbing individual BT comparisons affects both the estimated skills and
downstream leaderboard conclusions. Let $w\in\mathbb{R}^{N}$ denote match
weights and define the weighted BT estimator
\begin{equation}
    \hat{\theta}(w)
    =
    \operatorname*{arg\,min}_{\theta:\theta_1=0}
    \sum_{n=1}^{N} w_n \ell(z_n;\theta),
    \label{eq:weighted}
\end{equation}
where $w=\mathbf{1}$ gives the full-data fit and $w_n=0$ removes match $z_n$.
Let $\hat{\theta}=\hat{\theta}(\mathbf{1})$,
$g_n=\nabla_{\theta}\ell(z_n;\hat{\theta})$, and
$H=\nabla_{\theta}^{2}\sum_{n=1}^{N}\ell(z_n;\hat{\theta})$.
Influence functions approximate the local effect of changing a match weight
without refitting~\citep{koh_understanding_2020}:
\begin{equation}
    \frac{\partial \hat{\theta}(w)}{\partial w_n}\Big|_{w=\mathbf{1}}
    =
    -H^{-1}g_n .
    \label{eq:param_influence}
\end{equation}

We propagate this parameter sensitivity to a scalar leaderboard objective
$f(\hat{\theta}(w),w)$, such as a \TopkMem{}, \KTau{} surrogate, or
\CIUncert{} criterion. Allowing explicit dependence on $w$ captures objectives,
such as uncertainty, that change directly with the weighted comparison graph.
The objective-level influence of match $z_n$ is
\begin{equation}
\begin{gathered}
    I^{(f)}_n
    :=
    \nabla_{\theta} f(\hat{\theta}(w),w)^{\top}
    \frac{\partial \hat{\theta}(w)}{\partial w_n}
    +
    \frac{\partial f(\hat{\theta}(w),w)}{\partial w_n},
    \\[2pt]
    f(\hat{\theta}(w+\Delta w),w+\Delta w)
    \approx
    f(\hat{\theta}(w),w)
    +
    \sum_{n=1}^{N}\Delta w_n I^{(f)}_n .
\end{gathered}
\label{eq:propagate}
\end{equation}

In the leaderboard setting,
\citet{huang_dropping_2025} instantiate this idea with \TopkMem{} objectives
$f(\theta)=\theta_i-\theta_j$ across top-$k$ boundaries, showing that dropping a
small number of preference votes can change top-ranked membership. We generalize
this view by decoupling the perturbation mechanism from the audited objective:
the same propagation rule applies to multiple leaderboard objectives and, later,
to \Drop, \Add, and \Flip actions. Because our actions are finite rather than
infinitesimal, we follow \citep{huang_dropping_2025} use a one-step Newton (1sN) refinement to partially account
for curvature changes in the perturbed objective; details are in
Appendix~\ref{app:1sn}.
\section{Influence framework}
\label{sec:framework}

\begin{figure*}[t]
  \centering
  \includegraphics[width=0.95\textwidth,trim=0 0 0 10mm, clip]{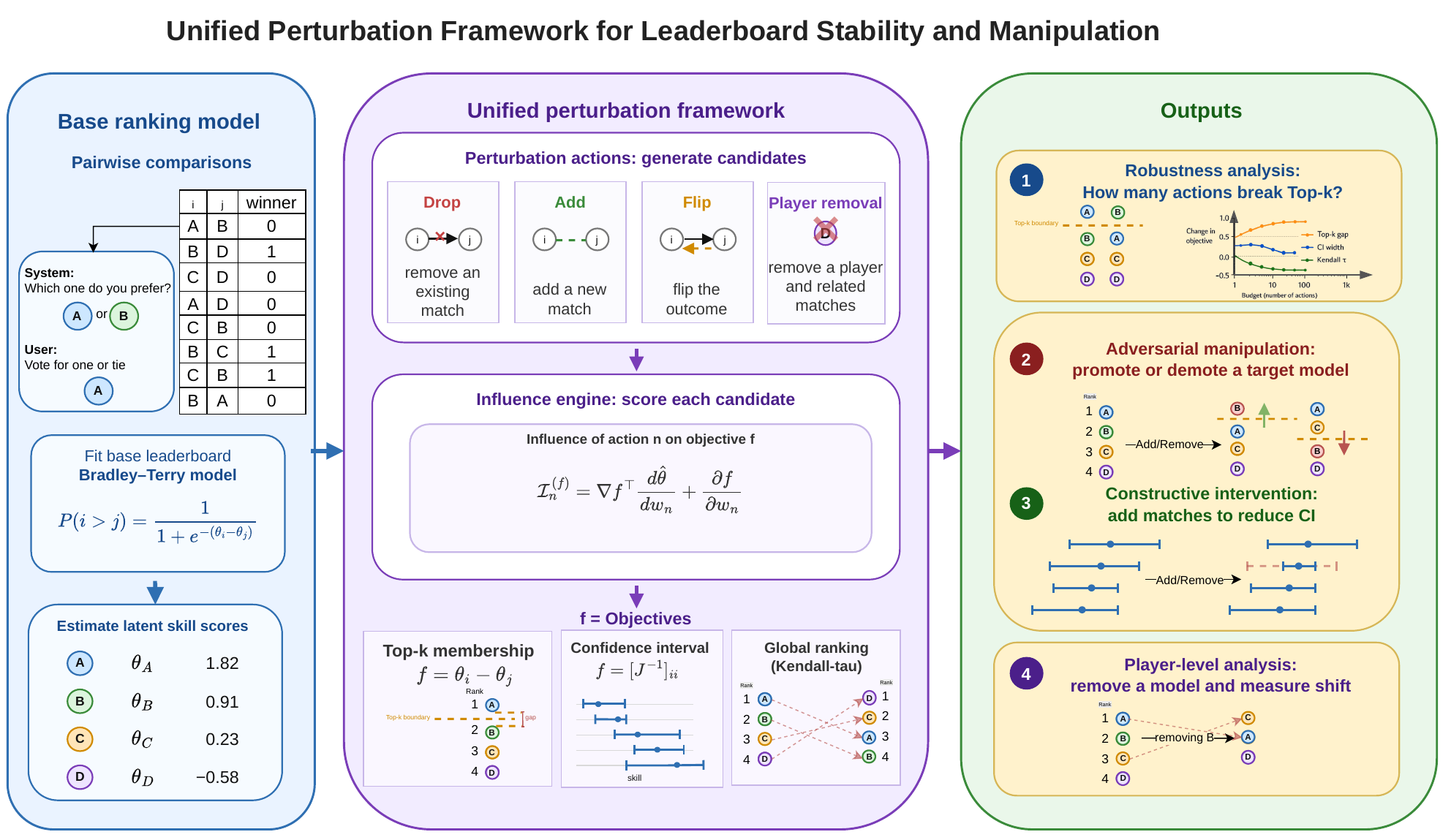}
  \caption{%
    Overview of the framework: influence scores are computed for each action
    type (\Drop, \Add, \Flip) and propagated through the Bradley--Terry
    estimator to ranking objectives (\TopkMem, \CIUncert, \KTau).
  }
  \label{fig:framework_overview}
\end{figure*}
\label{app:framework}
The goal of our framework is to systematically quantify how perturbations to the leaderboard dataset $D$ propagate to downstream ranking objectives, with an overview of the framework shown in Figure~\ref{fig:framework_overview}.

\subsection{Action space}
We consider three fundamental actions that characterize the ways a leaderboard can be manipulated. Each action corresponds to a specific modification of the weighted M-estimator defined in Eq.~\ref{eq:weighted}.

\textbf{Action 1: Match Dropping (\Drop).}
Following prior work that uses influence methods to show that LLM leaderboards are non-robust to small amounts of data removal~\citep{huang_dropping_2025}, we model dropping a match $z_n \in D$ as setting its weight from $w_n = 1$ to $w_n = 0$. The effect of this single perturbation on a downstream objective is approximated using influence as:
\begin{equation}
    \Delta f_{\text{drop}, n} \approx -\, \mathcal{I}_n^{(f)}.
\end{equation}

\textbf{Action 2: Match Addition (\Add).}
Adding a match involves selecting a candidate $z_{\text{new}}$ and increasing its weight from $w_{\text{new}} = 0$ toward $1$. Specifically, we define our candidate set as the union of the current dataset (where $w_n = 1$) and all possible pairs (where $w_n = 0$). We then identify which $w_{\text{new}}$ to shift from zero to one based on its estimated influence on the objective:
\begin{equation}
\begin{aligned}
    \Delta f_{\text{add}, \text{new}} &\approx (+1) \cdot \mathcal{I}_{\text{new}}^{(f)},
    \quad \text{where} \\
    \mathcal{I}_{\text{new}}^{(f)} &= - \nabla_\theta f^\top H^{-1} \nabla_\theta \ell(z_{\text{new}}; \hat{\theta}) \\
    &\quad + \left(\frac{\partial f}{\partial w_{\text{new}}}\right) .
\end{aligned}
\end{equation}
This action models different levels of control over the added data, from benign
data collection to stronger manipulation. We consider three candidate spaces of
increasing control:
\begin{itemize}[leftmargin=1.2em,noitemsep,topsep=0pt,parsep=0pt,partopsep=0pt]
    \item \textbf{all\_pairs:} Lowest control. The method selects an unordered 
    pair $(i,j)$, but the outcome is fixed by the estimated 
    skill ordering, modeling data collection without outcome control.

    \item \textbf{all\_outcomes\_weighted:} Intermediate control. Both outcomes 
($i \succ j$ and $j \succ i$) are considered separately, and each outcome's 
effect is weighted by its BT probability. These probability-weighted outcome 
scores are used for scoring and selection only.
    \item \textbf{all\_outcomes:} Strongest control. Both outcomes are treated as 
    separate candidates, allowing the method to select the outcome with largest 
    effect, modeling stronger manipulation.
\end{itemize}

\textbf{Action 3: Outcome Flipping (\Flip).}
Reversing the outcome of match $z_n$ to $z_n'$ (where $y_n' = 1 - y_n$) is equivalent to simultaneously dropping the original and adding the reversed match:
\begin{equation}
    \Delta f_{\text{flip}, n} \approx \mathcal{I}_{z_n'}^{(f)} - \mathcal{I}_{z_n}^{(f)}.
\end{equation}
We use \Flip primarily as a counterfactual robustness, explanation, and 
manipulation action, measuring how a leaderboard conclusion would change if an 
observed comparison had resolved in the opposite direction~\citep{pearl_causal_2009, wachter2018counterfactualexplanationsopeningblack}. This explains which 
outcomes most support the current ranking and which reversals would most 
effectively manipulate it.

\subsection{Objectives}
We characterize leaderboard robustness and sensitivity through three distinct classes of objectives. These scalar functionals, $f(\theta, w)$, allow us to quantify how data perturbations propagate to specific ranking outcomes, from local skill gaps to global ranking consistency and statistical uncertainty.

\textbf{Top-$k$ Membership Objective / Gap Objective (\TopkMem).}
The top-$k$ membership objective (gap objective), following prior work on leaderboard robustness \cite{huang_dropping_2025}, measures the difference in estimated skill between two models, which drives rank ordering and $\topk$ membership. For a target pair $(i, j)$, it is defined as: (the derivative is given in
Appendix~\ref{app:derivations})
\begin{equation}
f_{\text{gap}}(\theta) = \theta_i - \theta_j.
\label{eq:gap}
\end{equation}


\textbf{Statistical Uncertainty Objectives (\CIUncert).}
A confidence interval (CI) quantifies uncertainty around an estimated leaderboard
skill, indicating how precisely a model's latent strength is identified from the
comparison data. We use a local information approximation motivated by the BT
uncertainty analysis of~\cite{gao_uncertainty_2022}, where the
coordinate-wise uncertainty of the MLE is governed by the inverse local Fisher
information of each player.

For each unordered pair $(i,j)$, let
$p_{ij}=\sigma(\hat{\theta}_i-\hat{\theta}_j)$ and
$v_{ij}=p_{ij}(1-p_{ij})$.
Let
$w_{ij}=\sum_{n:\{i_n,j_n\}=\{i,j\}} w_n$
denote the total comparison weight assigned to the unordered pair $\{i,j\}$.
We define the local information of player $i$ as
$
    \hat{\rho}_i^2(w)
    =
    \sum_{j\neq i} w_{ij}v_{ij}.
    \label{eq:local_info_rho}
$
Intuitively, $\hat{\rho}_i^2(w)$ is large when player $i$ has many informative
comparisons, leading to smaller uncertainty. Following this local-information
view, we use its inverse as a scalable proxy for player-wise variance and define
\begin{equation}
    f_{\mathrm{CI\text{-}player}}(i;w)
    =
    \frac{1}{\hat{\rho}_i^2(w)},
    \qquad
    f_{\mathrm{CI\text{-}trace}}(w)
    =
    \sum_{i=1}^{M}\frac{1}{\hat{\rho}_i^2(w)}.
    \label{eq:rho_uncertainty_objectives}
\end{equation}
The first objective measures uncertainty for a target player, which we use for
targeted CI reduction, while the second provides a trace-style proxy for global
leaderboard uncertainty. These objectives depend both on $\hat{\theta}$ and explicitly on the comparison weights
$w$; their corresponding direct weight derivatives are given in
Appendix~\ref{app:derivations}. We discuss the relation between this Gao-style
local approximation and the more general sandwich covariance estimator in
Appendix~\ref{app:sandwich_gao_relation}.

\textbf{Global Ranking Consistency (\KTau).}
To evaluate how perturbations affect the entire leaderboard structure, we use a smooth surrogate of \KTau{} correlation against a reference ranking $\pi$ (which is the ranking from BT fitted on the whole initial dataset). Since the discrete \KTau{} is non-differentiable, we employ a $\tanh$-based relaxation:
\begin{equation}
f_{\tau,T}(\theta; \pi) = \frac{2}{M(M-1)} \sum_{a<b} s_{ab} \tanh\left(\frac{\theta_a - \theta_b}{T}\right),
\end{equation}
where $s_{ab} \in \{-1, +1\}$ encodes the relative order of items $a$ and $b$ in the reference $\pi$, and $T$ is a temperature parameter. As $T \to 0$, this objective converges to the discrete \KTau. This differentiable form allows us to compute influence scores $\mathcal{I}_n^{(\tau,T)}$ to identify which specific comparisons most heavily affect overall ranking consistency. The derivative is given in
Appendix~\ref{app:derivations}.

\subsection{Player-level influence}
Following~\citet{giordano_swiss_nodate}, we extend influence functions from
individual matches to structured group perturbations. For a subset
$\mathcal{G}\subset\{1,\dots,N\}$ of matches, such as all matches incident to a
given player, the first-order change in the BT estimator is
\begin{equation}
\begin{aligned}
\Delta \theta_{\mathcal{G}}
&\approx
\sum_{n \in \mathcal{G}} \Delta w_n
\frac{\partial \hat{\theta}(w)}{\partial w_n} \\
&=
- H(w)^{-1}
\sum_{n \in \mathcal{G}} \Delta w_n\,
\nabla_\theta \ell(z_n;\hat{\theta}(w)).
\end{aligned}
\label{eq:group_influence}
\end{equation}
We then propagate $\Delta\theta_{\mathcal{G}}$ through a scalar objective
$f(\theta,w)$ as in Eq.~\ref{eq:propagate}, yielding a player-level influence
score from the aggregate contribution of that player's incident matches.

For player-removal analysis, we additionally use a grouped Newton refinement:
after the first-order grouped deletion step in Eq.~\ref{eq:group_influence}, we
take one Newton correction on the remaining comparison data. This is analogous
to the 1sN correction for single-match perturbations, but applied jointly to all
matches incident to the removed player; details are in
Appendix~\ref{app:group-newton}.

\section{Experiments}


\paragraph{Dataset.} 
Our primary evaluation uses {Arena-55K}~\cite{lmarena_arena55k} (64 LLMs, 55k human votes), a subset of Chatbot Arena~\cite{chiang2024chatbot}. We additionally evaluate on six pairwise-comparison datasets: Chatbot Arena LLM Judges and MT-Bench Human Evaluation~\cite{zheng2023judging}, Vision Arena~\cite{chou_visionarena_2025}, WebDev Arena~\cite{vichare_webdev_2025}, NBA Elo Top-50~\cite{fivethirtyeight_nba_elo_2025}, and ATP Tennis Top-10~\cite{sackmann_tennis_atp_2024}. Dataset details are in Appendix~\ref{app:dataset-descriptions}.

\paragraph{Experimental protocol.}
Unless otherwise stated, we greedily select perturbations using influence scores
under a fixed budget, without refitting during selection. We refit after the
selected perturbations to report the exact post-perturbation leaderboard and
success conditions, though these checks could be monitored by influence estimates.
Full algorithms are provided in Appendix~\ref{app:algorithms}.

\subsection{Match analysis}
\label{sec:match-analysis}

We study how individual matches affect leaderboard outcomes along two axes: (i) a {\textit{robustness}} question: how many perturbations are needed before a criterion changes?, and (ii) a {\textit{manipulation}} question: can we steer ranking or confidence towards a desired outcome?

\subsubsection{Robustness analysis}
\label{sec:robustness}

\paragraph{\textit{Top-k membership} robustness}
\label{par:topk-robustness}
Following \citet{huang_dropping_2025}, we measure the minimum number of match 
perturbations required to change the composition of the $\topk$ set. We 
instantiate our framework with the gap objective $f_{\text{gap}}$ and score all actions candidates by their 
predicted effect on crossing a top-$k$ boundary. Starting from the original 
leaderboard, we greedily apply the highest-influence action until the 
$\topk$ set changes or until reaching a budget of 5\% of the dataset is reached.

Table~\ref{tab:actions_summary} reports the minimum influence-guided actions
needed to change the Top-1 model. On major LLM leaderboards, fewer than 1\%
targeted actions suffice (e.g., 3 \Flip{} or 5 \Drop{} on Arena 55k). Across
datasets, \Flip{} is the most action-efficient perturbation, follows by \Drop{} while \Add{}
variants typically require larger budgets or remain robust within the budget.
We ablate the
effect of varying $k$ on Arena 55k in Appendix~\ref{app:budget-k}.
\begin{table*}[t]
  \vspace{-0.3em}
  \centering
  \footnotesize
  \setlength{\tabcolsep}{5pt}
  \caption{%
Minimum influence-guided actions needed to change the Top-1 model under a 5\%
budget. Percentages are relative to total matches; {robust} means no change
within budget.
}
  \resizebox{0.8\textwidth}{!}{%
  \begin{tabular}{l r r r r r r}
  \toprule
  \textbf{Dataset} & \textbf{Dataset size} & \textbf{Drop} & \textbf{Flip} & \textbf{Add-pairs} & \textbf{Add-outcomes} & \textbf{Add-weighted} \\
  \midrule
  Arena 55k       & 57{,}477  & 5 (0.01\%)   & 3 (0.01\%)   & 9 (0.02\%)   & 9 (0.02\%)   & 9 (0.02\%)   \\
  Arena LLM-J     & 49{,}938  & 9 (0.02\%)   & 6 (0.01\%)   & 22 (0.04\%)  & 21 (0.04\%)  & 21 (0.04\%)  \\
  MT-Bench        & 3{,}355   & 92 (2.74\%)  & 46 (1.37\%)  & robust        & robust        & robust        \\
  NBA Top-50      & 109{,}892 & 24 (0.02\%)  & 15 (0.01\%)  & 77 (0.07\%)  & 55 (0.05\%)  & 57 (0.05\%)  \\
  ATP Top-10      & 278       & 6 (2.16\%)   & 3 (1.08\%)   & robust        & 9 (3.24\%)   & 14 (5.00\%)  \\
  Vision Arena    & 29{,}849  & 50 (0.17\%)  & 25 (0.08\%)  & robust        & robust        & robust        \\
  WebDev Arena    & 10{,}501  & 179 (1.70\%) & 12 (0.11\%)  & robust        & robust        & robust        \\
  \bottomrule
  \end{tabular}
  }
  \label{tab:actions_summary}
  \vspace{-0.5em}
  \end{table*}

\paragraph{CI-aware \textit{Top-$k$ membership} robustness}
We extend the Top-$k$ robustness analysis to incorporate estimation uncertainty.
For a chosen boundary rank $k$, we augment the usual
membership objective with confidence bounds: we fix the boundary pair consisting
of the model currently ranked $k$ and the model currently ranked $k+1$, and
compare the upper confidence bound of the rank-$k$ model to the lower confidence
bound of the rank-$(k+1)$ model. A perturbation is declared successful only if,
after refitting, the rank-$(k+1)$ model is not merely above the rank-$k$ model in
point estimate, but is \emph{strictly} above under uncertainty as well, meaning
that its lower CI bound exceeds the rank-$k$ model's upper CI bound. This
strict CI-aware criterion is therefore stronger than an ordinary non-CI-aware (point-estimate)
Top-$k$ membership; the algorithm and an example are given in
Appendix~\ref{app:algorithms}, ~\ref{sec:topk-ci-vs-nonci}.

Table~\ref{tab:ci_vs_nonci_actions} compares the minimum number of 
influence-guided actions needed under point-estimate and CI-aware criteria, using per dataset boundary rank $k$ for each dataset. Accounting for uncertainty 
increases the required budget, but does not make the leaderboard robust.  
CI-aware criteria require on average $13\times$ more actions than 
point-estimate changes, yet several datasets remain non-robust within the 
allowed budget. \Drop and \Flip remain the most effective actions, while \Add 
variants often require larger budgets or remain robust within the budget.
\begin{table*}[t]
    \vspace{-0.3em}
    \centering
    \caption{
      Minimum number of actions needed to change the top-$k$ boundary under
{point-estimate / CI-aware} criteria (percentages relative to total matches) with
95\% confidence level and a 5\% perturbation budget. Across datasets, the
CI-aware criterion generally requires substantially more actions than the
point-estimate criterion. {robust}: target not changed within the 5\% budget.
    }
    \setlength{\tabcolsep}{3pt}
    \resizebox{\textwidth}{!}{%
    \begin{tabular}{l r r r r r r r}
    \toprule
    \textbf{Dataset} & \textbf{Dataset size} & \textbf{$k$} & \textbf{Drop (point-estimate / CI-aware)} & \textbf{Flip (point-estimate / CI-aware)} &
  \textbf{Add-pairs} & \textbf{Add-outcomes} & \textbf{Add-weighted} \\
    \midrule
    Arena 55k        & 57{,}477  & 22 & 2 (0.00\%) / 19 (0.03\%) & 1 (0.00\%) / 12
  (0.02\%) & 2 (0.00\%) / 39 (0.07\%) & 2 (0.00\%) / 26 (0.05\%) & 3 (0.01\%) / 28
  (0.05\%) \\
    LLM Judge Arena  & 49{,}938  & 31 & 1 (0.00\%) / 31 (0.06\%) & 1 (0.00\%) / 17
  (0.03\%) & 1 (0.00\%) / 50 (0.10\%) & 1 (0.00\%) / 32 (0.06\%) & 1 (0.00\%) / 45
  (0.09\%) \\
    MT-Bench         & 3{,}355   & 2  & 13 (0.39\%) / 62 (1.85\%) & 7 (0.21\%) / 33
  (0.98\%) & robust / robust & robust / robust & robust / robust \\
    NBA Top-50       & 109{,}892 & 8  & 2 (0.00\%) / 46 (0.04\%) & 1 (0.00\%) / 25
  (0.02\%) & 5 (0.00\%) / robust & 3 (0.00\%) / robust & 6 (0.01\%) / robust \\
    ATP Top-10       & 278       & 8  & 1 (0.36\%) / robust & 1 (0.36\%) / 7 (2.52\%)
  & 1 (0.36\%) / robust & 1 (0.36\%) / robust & 1 (0.36\%) / robust \\
    Vision Arena     & 29{,}849  & 14 & 4 (0.01\%) / 43 (0.14\%) & 3 (0.01\%) / 33
  (0.11\%) & 6 (0.02\%) / robust & 5 (0.02\%) / robust & 10 (0.03\%) / robust \\
    WebDev Arena     & 10{,}501  & 12 & 3 (0.03\%) / 342 (3.26\%) & 2 (0.02\%) / 23
  (0.22\%) & 7 (0.07\%) / robust & 6 (0.06\%) / robust & 10 (0.10\%) / robust \\
    \bottomrule
    \end{tabular}
    }
    \label{tab:ci_vs_nonci_actions}
    \vspace{-0.5em}
  \end{table*}



\begin{figure*}[t]
  \vspace{-0.3em}
  \centering
  
  \begin{subfigure}{0.45\linewidth}
    \centering
    \includegraphics[width=\linewidth]{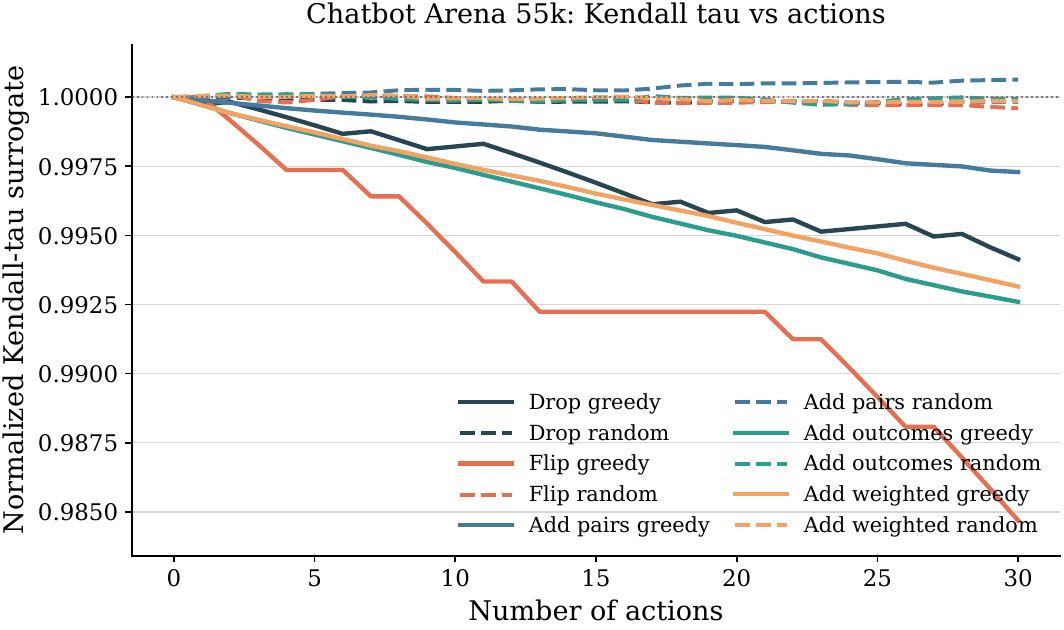}
    \caption{Global ranking robustness (\KTau).}
    \label{fig:tau_curve}
  \end{subfigure}
  \hfill
  \begin{subfigure}{0.45\linewidth}
    \centering
    \includegraphics[width=\linewidth]{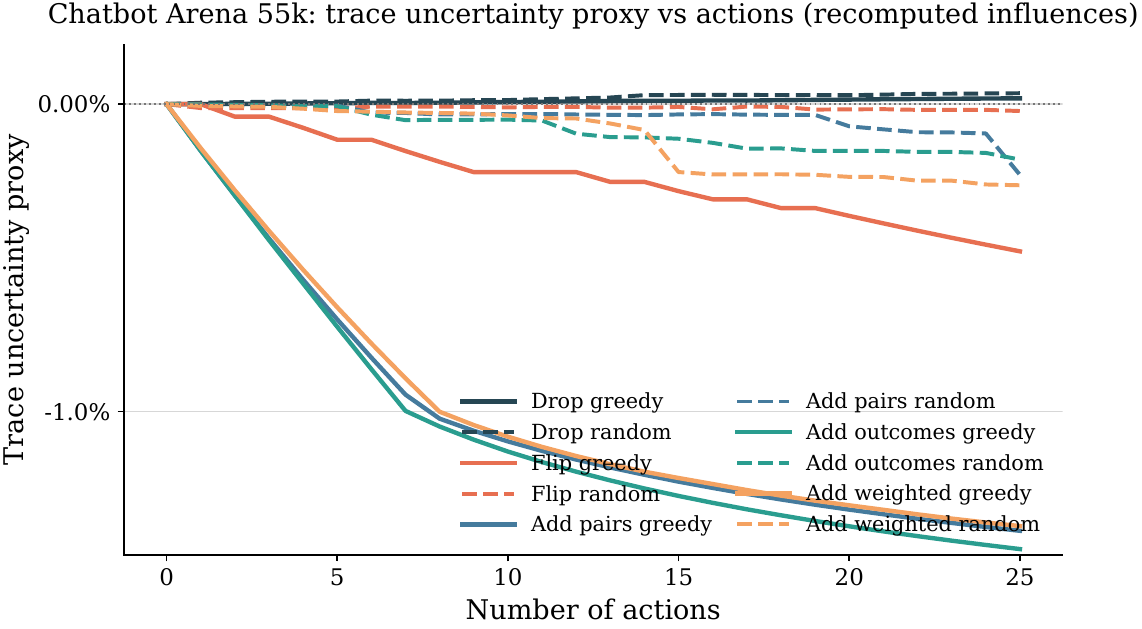}
    \caption{Trace uncertainty change (\%).}
    \label{fig:trace_curve}
  \end{subfigure}
  
  \caption{
  Global ranking and uncertainty robustness on Chatbot Arena 55k under budgets of
30 actions for global ranking and 25 for uncertainty (solid = influence-guided,
dashed = random). \textbf{Left:} Influence-guided \Flip causes the largest
degradation ($\tau \approx 0.985$), while \Drop is more conservative
($\tau \approx 0.994$); random baselines remain near 1.0. \textbf{Right:}
Influence-guided \Add reduces trace uncertainty by up to $\approx -1\%$, whereas
random actions slightly increase it.
  }
  
  \label{fig:global_ci_curves}
  \vspace{-0.5em}
\end{figure*}

\paragraph{Global ranking robustness (\KTau).}
\label{par:tau-robustness}
Beyond local boundary changes, we score candidates using the $f_\tau$ surrogate and apply actions greedily, comparing against random selection (Figure~\ref{fig:global_ci_curves}, left). Influence-guided \Flip{} causes the steepest $\tau$ degradation, while \Drop{} is noticeably more conservative and \Add{} variants cluster at intermediate degradation. In all cases, random selection barely moves $\tau$, confirming that influence-guided selection is essential, random perturbations are insufficient to disrupt global ranking even at budget 30. Results for all datasets are in Appendix~\ref{app:tau-trace-curves}.

\paragraph{\textit{Confidence-interval-based uncertainty} robustness.}
\label{par:ci-robustness}
We next evaluate how perturbations affect global leaderboard uncertainty. Using
$f_{\text{CI-trace}}$ to score candidates and applying actions greedily
(Figure~\ref{fig:global_ci_curves}, right), the action-type ordering reverses:
\Add{} variants yield the steepest uncertainty reduction, reaching roughly
$-1\%$ of the initial trace by budget 25, about an order of magnitude larger
than \Drop{} or \Flip{}. In contrast, random additions slightly increase
uncertainty, as they tend to introduce less-informative comparisons. Results for
all datasets are in Appendix~\ref{app:tau-trace-curves}.

\paragraph{From influence scores to leaderboard audits}
\label{sec:influence-audits}

\begin{figure*}[t]
    \vspace{-0.3em}
    \centering

    \begin{minipage}[t]{0.49\linewidth}
        \vspace{0pt}
        \centering
        \includegraphics[width=\linewidth]{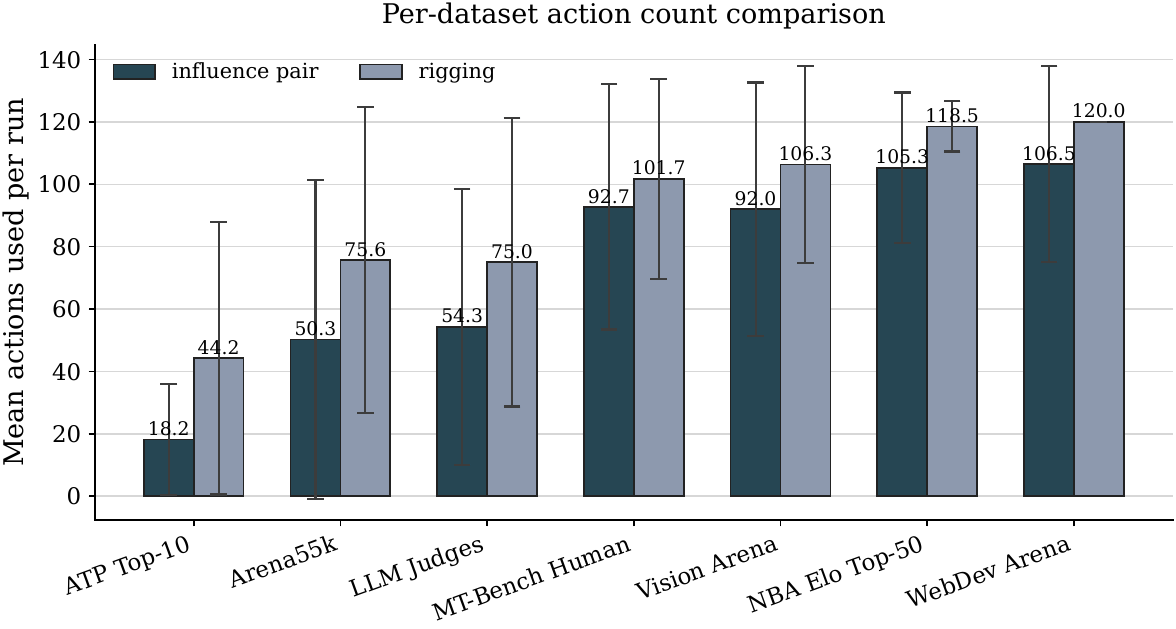}
        \captionof{figure}{%
        Mean intervention actions per run for pairwise influence versus the rigging baseline.
        Error bars show variability across repeated trials. Each run is capped at 120 exposed pairs.
        }
        \label{fig:rigging_vs_ours}
    \end{minipage}
    \hfill
    \begin{minipage}[t]{0.49\textwidth}
        \vspace{0pt}
        \centering
        \captionof{table}{%
        Dataset-level robustness scores across the Top-1, CI views, and global ranking.
        Lower values indicate less robustness under a 5\% data budget.
        \textbf{Bold} marks the most robust dataset.
        }
        \label{tab:dataset-robustness-scores}
        \scriptsize
        \renewcommand{\arraystretch}{1.08}
        \resizebox{\linewidth}{!}{%
        \begin{tabular}{lrrrr}
        \toprule
        \textbf{Dataset} &
        $\boldsymbol{R_{\mathrm{Top}\text{-}1}}$ &
        $\boldsymbol{R_{\mathrm{CI}}}$ &
        $\boldsymbol{R_{\tau}}$ &
        $\boldsymbol{R_{\mathrm{all}}}$ \\
        \midrule
        Arena 55k     & 0.001 & 0.976 & \textbf{0.993} & 0.656 \\
        Arena LLM-J   & 0.000 & 0.975 & 0.992 & 0.656 \\
        MT-Bench      & 0.048 & 0.955 & 0.985 & \textbf{0.662} \\
        NBA Top-50    & 0.003 & \textbf{0.992} & 0.982 & 0.659 \\
        ATP Top-10    & \textbf{0.072} & 0.810 & 0.474 & 0.452 \\
        Vision Arena  & 0.003 & 0.976 & 0.991 & 0.657 \\
        WebDev Arena  & 0.004 & 0.579 & 0.782 & 0.455 \\
        \bottomrule
        \end{tabular}
        }
    \end{minipage}

    \vspace{-0.5em}
\end{figure*}

Beyond identifying non-robustness, we define normalized dataset-level robustness
metrics that summarize influence-guided failures as audit signals. These metrics
measure either the budget needed to change a leaderboard conclusion, as in
Top-1, or the normalized magnitude of change, as in
Kendall-$\tau$ and confidence interval stability, Appendix~\ref{app:robustness-scores} gives the full metric
definitions. Table~\ref{tab:dataset-robustness-scores} reports
these scores for all seven dataset. Lower values indicate less robustness. Based on
$R_{\mathrm{all}}$, MT-Bench is the most robust dataset, followed closely by
NBA Top-50 and the arena-style LLM datasets, whereas ATP Top-10 and WebDev
Arena are least robust overall due to lower global-ranking stability.


\paragraph{Top-$k$ entry and removal.}
\label{par:topk-manipulation}

We use Top-$k$ entry and removal as targeted manipulation tasks: promotion moves
an outside model into the Top-$k$ set, while demotion pushes an inside model out.
We evaluate both under a sequentially revealed stream of candidate comparisons.
For each dataset, we partition the leaderboard into top, middle, and lower skill
regions and sample one boundary rank from each region; promotion targets rank
$(k+1)$ and demotion targets rank $k$. We compare our influence-guided policy
with an omni-rigging baseline~\citep{min_improving_2025} under the same exposed
stream, where each method may discard a pair, accept either directional outcome,
or encode a tie; full details are in Appendix~\ref{app:algorithms}. Our method
uses a dynamic boundary-gap objective, recomputing the boundary opponent after
each accepted perturbation and selecting the action with largest predicted
influence on the current gap.

Figure~\ref{fig:rigging_vs_ours} shows that the influence-guided policy typically achieves targeted promotion and demotion with fewer actions than the omni-rigging baseline across most datasets. Across datasets, influence-guided selection uses about 74 actions on average, compared with about 92 for omni-rigging, a roughly 19\% reduction. This indicates that directly optimizing the current $\topk$ boundary gap is more efficient than locally greedy rank manipulation.

\paragraph{CI reduction for a target model}
\label{par:ci-targeted}
From a leaderboard provider's perspective, we study whether the framework can
constructively reduce the confidence interval of a target model $m^{*}$. We use
the target uncertainty objective $f_{\mathrm{CI\text{-}player}}(m^{*})$ and
restrict the action space to \Add, selecting candidate additions with the most
negative predicted influence. We compare influence-guided \Add with the
\texttt{all\_pairs} candidate space against \emph{Random} and \emph{Arena
Active}~\citep{chiang_chatbot_2024}; Random samples pairs uniformly, while Arena
Active prioritizes target-involving uncertain comparisons. After selecting a
pair, all methods assign the outcome by the current BT skill ordering, refit
after each addition, and measure target-model CI reduction.

For each dataset, we choose three target players from the top, middle, and lower
skill regions and run each method for 12 \Add{} actions. Table~\ref{tab:ci}
reports $\% \Delta \mathrm{CI}=100\times(\mathrm{final}-\mathrm{initial})/\mathrm{initial}$,
averaged over targets; more negative values indicate larger uncertainty
reduction. Averaged over all results, influence-guided sampling produces a larger CI-width reduction than \textit{ Arena Active} and \textit{Random}, showing directly optimizing the
target CI objective is more informative than Arena-style
heuristics.

\subsection{Player removal analysis}
\label{sec:player-removal}

Beyond individual match perturbations, we consider {player removal}: 
dropping all matches of a given model, as occurs when models are retired, 
deprecated, or excluded from evaluation. Prior work on the leaderboard 
illusion~\citep{singh_leaderboard_2025} shows that excluding models can 
qualitatively change which remaining models appear superior; we quantify this 
effect systematically. We use our influence framework to rank all players by predicted 
$|\tau|$-influence and validate by removing the most influential player 
in each dataset and measuring the resulting \KTau{} shift. 
Table~\ref{tab:mip_ablation_summary} shows that removing an influential 
player can induce broad reordering, moving up to 28 players, shifting ranks 
up to 13 positions, and changing up to 6 top-10 memberships.
Connectivity is the strongest and most consistent predictor 
(Appendix~\ref{app:player-feature-association}).

\begin{table*}[t]
\centering

\begin{minipage}[t]{0.48\linewidth}
    \vspace{0pt}
    \centering
    \captionof{table}{%
    Percent change in target-model CI width after 12 targeted actions,
$\% \Delta \mathrm{CI}=100\times(\mathrm{final}-\mathrm{initial})/\mathrm{initial}$. More negative is better.
    }
    \label{tab:ci}
    \vspace{2pt}
    \scriptsize
    \renewcommand{\arraystretch}{1.08}
    \setlength{\tabcolsep}{3pt}
    \resizebox{\linewidth}{!}{%
    \begin{tabular}{lrrr}
    \toprule
    Dataset & Influence-based & Arena Active & Random \\
    \midrule
    
    MT-Bench     & \textbf{-0.27 $\pm$ 0.05} & {-0.25 $\pm$ 0.07} & -0.08 $\pm$ 0.03 \\
    WebDev Arena & -0.28 $\pm$ 0.05 & \textbf{-0.39 $\pm$ 0.23} & -0.02 $\pm$ 0.00 \\
    Vision Arena & \textbf{-0.17 $\pm$ 0.12} & {-0.16 $\pm$ 0.11} & -0.01 $\pm$ 0.01 \\
    Arena LLM-J  & \textbf{-0.47 $\pm$ 0.25} & -0.30 $\pm$ 0.11 & -0.00 $\pm$ 0.00 \\
    Arena 55k    & \textbf{-0.72 $\pm$ 0.22} & -0.54 $\pm$ 0.28 & -0.01 $\pm$ 0.01 \\
    NBA Top-50   & \textbf{-0.33 $\pm$ 0.09} & -0.29 $\pm$ 0.05 & -0.01 $\pm$ 0.00 \\
    \textbf{All datasets} & \textbf{-0.37 $\pm$ 0.13} & -0.32 $\pm$ 0.14 & -0.02 $\pm$ 0.01 \\
    \bottomrule
    \end{tabular}
    }
\end{minipage}
\hfill
\begin{minipage}[t]{0.49\linewidth}
    \vspace{0pt}
    \centering
    \captionof{table}{%
    Most-influential-player ablation results. We remove the player with largest
    influence and all associated matches, then measure changes.%
    }
    \label{tab:mip_ablation_summary}
    \vspace{2pt}
    \scriptsize
    \renewcommand{\arraystretch}{1.08}
    \setlength{\tabcolsep}{3pt}
    \resizebox{\linewidth}{!}{%
    \begin{tabular}{lrrrrr}
    \toprule
    \textbf{Dataset} & $\boldsymbol{\Delta\tau}$ & \textbf{Moved} & \textbf{Max Shift} & \textbf{Top-10} & \textbf{Removed \%} \\
    \midrule
    ATP Top-10   & -0.167 & 2  & 2  & 2 & 24.1\% \\
    MT-Bench     &  0.000 & 0  & 0  & 0 & 30.3\% \\
    WebDev       & -0.091 & 5  & 2  & 4 & 23.2\% \\
    Vision       & -0.050 & 2  & 2  & 0 & 14.1\% \\
    Arena LLM-J  & -0.023 & 25 & 5  & 4 & 6.2\% \\
    Arena 55k    & -0.030 & 28 & 10 & 5 & 6.2\% \\
    NBA Top-50   & -0.053 & 27 & 13 & 6 & 0.5\% \\
    \bottomrule
    \end{tabular}
    }
\end{minipage}

\vspace{-0.5em}
\end{table*}

\section{Related work}
\label{sec:related-work}

\textbf{Ranking systems and evaluation leaderboards.}
The Bradley--Terry model~\citep{bradley_rank_1952, hamilton_many_2025, fang_recent_2026} and related models like
Elo rating systems~\citep{boubdir_elo_nodate, liu_am-elo_2025}, form the basis of modern probabilistic ranking. Maximum-likelihood estimation and uncertainty quantification for BT models are well-studied~\citep{hunter_mm_2004, bong_generalized_nodate, gao_uncertainty_2022, fan_uncertainty_2024}, with spectral and sorting-based methods providing efficient alternatives~\citep{negahban_rank_2015, wauthier_efficient_2013}. In LLM evaluation, pairwise preference aggregation has become central through Chatbot Arena~\citep{chiang_chatbot_2024}, with recent work improving statistical reliability~\citep{ameli_statistical_2025, gao_re-evaluating_2025} and studying evaluation bias~\citep{daynauth_ranking_2025, levtsov_confidence_2025}.

\textbf{Leaderboard robustness and manipulation.}
Benchmark rankings can be sensitive to dataset shifts, benchmark overfitting, and small comparison perturbations~\citep{boubdir_elo_nodate, huang_dropping_2025, singh_leaderboard_2025}. The leaderboard illusion~\citep{singh_leaderboard_2025}shows that benchmark
rankings can be distorted by selective disclosure, private testing, uneven data
allocation, and model inclusion or exclusion decisions; while vote manipulation and adversarial leaderboard attacks expose additional manipulation risks~\citep{min_improving_2025, huang_exploring_nodate, suri_exploiting_2025}. Related theory studies robustness of estimators to sample removal~\citep{azar_robustness_2025, broderick_automatic_2023}, and recent uncertainty-aware ranking methods propose rank sets or confidence diagrams~\citep{chatzi_prediction-powered_nodate, wang_confidence_2025}. Our work provides a unified framework covering match addition, removal, and flipping, extending prior work that typically studies one \PertType{} or one objective.

\textbf{Explainability and data attribution.}
Data attribution is a central tool for explaining model behavior by assigning
importance scores to individual training examples or groups of examples. Existing
approaches include Data Shapley~\citep{ghorbani_data_2019}, TracIn~\citep{pruthi_estimating_2020},
and TRAK~\citep{park_trak_2023}, which quantify how training data contributes to
model predictions or learned representations; see \citet{hammoudeh_training_2024}
for a survey. Influence functions are another classical approach to data
attribution, estimating the effect of training points on model parameters without
retraining~\citep{koh_understanding_2020, pregibon_logistic_1981}, with roots in
robust statistics~\citep{huber_behavior_1967, freedman_so-called_2006}.
Extensions include infinitesimal jackknife~\citep{giordano_swiss_nodate},
group influence methods~\citep{koh_accuracy_2019}, second-order
approximations~\citep{basu_second-order_2020}, and analyses of when influence
estimates are reliable~\citep{bae_if_nodate}. We build on this line of work by
propagating influence scores to leaderboard-specific objectives.

\section{Conclusion, limitations, and future work}

We introduced a unified influence-based perturbation framework for evaluating
Bradley--Terry leaderboard stability and manipulation. Across seven datasets, we
show that leaderboard conclusions are fragile: Top-1 membership changes
with fewer than 1\% targeted actions on major LLM leaderboards, CI-aware
Top-$k$ changes require larger budgets but remain vulnerable, and removing
highly connected players can induce broad rank shifts. Our normalized
dataset-level robustness scores summarize these effects across Top-$k$,
CI-aware, and global-ranking views. Overall, expert-curated MT-Bench is generally
more robust, while crowdsourced arena-style leaderboards are more sensitive;
among actions, \Flip is usually most effective for changing rankings, whereas
\Add is most useful for reducing uncertainty through targeted data collection.
Beyond auditing, the same framework enables targeted manipulation: influence-guided
actions promote or demote selected models with fewer interventions than the
omni-rigging baseline, while supporting constructive uncertainty reduction through comparisons that
narrow confidence intervals more effectively than Arena-style sampling.

Our analysis has several limitations. Influence estimates are local approximations and may be less accurate under large perturbation budgets or highly nonlinear ranking changes. Our uncertainty objectives use scalable BT-based proxies, which do not capture all sources of annotation noise, judge bias, prompt-level dependence, or model-specific evaluation artifacts. Future work should study higher-order attribution methods, outlier-robust BT estimators, and sample-complexity guarantees for leaderboard stability. Another important direction is to better understand what makes a match, player, or dataset highly influential or robust. Finally, the framework can be extended as a defensive tool, where low robustness
scores or highly influential comparisons trigger conservative reporting, targeted
data collection, or delayed updates, rather than being treated as evidence of
invalid data.



\section*{Impact Statement}

This paper studies the robustness and manipulability of pairwise-comparison
leaderboards used for benchmarking large language models and other ranked
systems. Our findings show that widely used leaderboards can be perturbed with
small, targeted modifications, which has dual-use implications: the same
techniques that enable auditing for fragility could in principle be used to
manipulate rankings. We highlight this tension explicitly and frame the
framework as primarily an auditing and defensive tool, where low robustness
scores or highly influential comparisons can motivate conservative reporting,
targeted data collection, or delayed updates. More broadly, we believe that
exposing these vulnerabilities promotes more reliable evaluation protocols and
better-calibrated trust in benchmark-driven model adoption decisions. There are
many potential societal consequences of advancing the field of Machine
Learning, none which we feel must be specifically highlighted here beyond the
considerations above.

\bibliographystyle{icml2026}
\bibliography{active}

\newpage
\appendix
\onecolumn

\section{Datasets}\label{app:datasets}

\subsection{Dataset descriptions}\label{app:dataset-descriptions}

We evaluate our framework on seven pairwise comparison datasets spanning LLM evaluation and sports ranking.

\paragraph{Chatbot Arena 55k.}
A crowdsourced platform where users simultaneously interact with two anonymous chatbots and vote for the preferred response~\cite{chiang_chatbot_2024}. We use the \texttt{arena-human-preference-55k} split, containing 57{,}477 human preference judgements across 64 models. Its large scale and diverse user base make it the primary benchmark for LLM evaluation, but also expose it to noise and adversarial risk.

\paragraph{Chatbot Arena LLM Judges.}
A companion dataset from the same platform in which pairwise preferences are collected using an LLM-as-a-judge rather than human votes~\cite{chiang_chatbot_2024}. The \texttt{chatbot-arena-llm-judges} split contains 49{,}938 comparisons across 64 models, allowing us to contrast automated and human evaluation robustness.

\paragraph{MT-Bench Human Judgments.}
A curated multi-turn benchmark designed to evaluate instruction-following and reasoning~\cite{chiang_chatbot_2024}. Preferences were collected from 58 expert-level annotators (predominantly graduate students), yielding 3{,}355 high-quality pairwise judgements. Its smaller size and expert annotations make it substantially more robust than crowdsourced platforms.

\paragraph{Vision Arena.}
A crowdsourced arena for vision-language models in which users compare two anonymous models on visual question-answering tasks. We use the \texttt{lmarena-ai/VisionArena-Battle} dataset, which contains 29{,}849 single- and multi-turn conversations.

\paragraph{WebDev Arena.}
A crowdsourced arena focused on web-development tasks such as building interactive applications and webpages. We use the \texttt{lmarena-ai/webdev-arena-preference-10k} dataset (10{,}501 prompts), which provides domain-specific evaluation complementary to open-ended chat.

\paragraph{ATP Top-10 Tennis.}
Match records from the ATP tour (2020--2024). We restrict to the top-10 ranked players by 2024 season standing who each played at least 20 matches, yielding 278 games in total. This small, sparse graph tests our framework in a regime where each individual match carries high weight.

\paragraph{NBA Elo Top-50.}
Historical NBA game records from all seasons. We focus on the top-50 teams by total games played, yielding 109{,}892 matchups. Its large, dense comparison graph represents the opposite extreme from ATP and assesses robustness in high-data settings.

\subsection{Dataset processing}
\label{app:data-processing}

Each pairwise comparison in the raw datasets is represented as a directed observation $(i, j, y)$ where $y \in \{0, 1\}$ indicates whether item $i$ beat item $j$. To handle ties uniformly, we adopt a symmetric formulation: each tied comparison is decomposed into two directed observations---$(i, j, 1)$ and $(j, i, 1)$---treating both items as winning once against the other.

For consistency, each non-tied comparison is also represented as two directed observations: $(i, j, y)$ and $(j, i, 1-y)$. This ensures a uniform representation across all comparisons and simplifies the construction of the feature matrix $X \in \mathbb{R}^{N \times M}$ used in the BT likelihood, where each row $x_n = e_i - e_j$ encodes the matched pair. The resulting matrix $X$ and outcome vector $y$ are the direct inputs to the weighted M-estimator in Eq.~\ref{eq:weighted}.


\section{Framework details}\label{app:framework-fig}

\subsection{BT input--output format and tie handling}
\label{app:bt-io-format}

The Bradley--Terry model takes a directed comparison dataset as input and
outputs a latent skill vector for the ranked players. Let $M$ denote the number
of players and let $e_i \in \mathbb{R}^M$ be the standard basis vector for
player $i$. Each directed comparison between players $i$ and $j$ is encoded as
\[
    x_{ij} = e_i - e_j,
\]
where the binary outcome $y_{ij} \in \{0,1\}$ indicates whether the first player
in the directed pair wins. After preprocessing, the full BT input is represented
as
\[
    X =
    \begin{bmatrix}
    x_1^\top \\
    x_2^\top \\
    \vdots \\
    x_N^\top
    \end{bmatrix}
    \in \{-1,0,1\}^{N \times M},
    \qquad
    y =
    \begin{bmatrix}
    y_1 \\
    y_2 \\
    \vdots \\
    y_N
    \end{bmatrix}
    \in \{0,1\}^{N},
\]
where each row $x_n=e_{i_n}-e_{j_n}$ contains one $+1$ entry for the first
player, one $-1$ entry for the second player, and zeros elsewhere.

Given this input, the BT model estimates a skill vector
\[
    \hat{\theta}
    =
    \begin{bmatrix}
    \hat{\theta}_1 \\
    \hat{\theta}_2 \\
    \vdots \\
    \hat{\theta}_M
    \end{bmatrix}
    \in \mathbb{R}^{M},
\]
where larger values indicate stronger players. The main BT outputs are the
fitted skill vector $\hat{\theta}$, the induced pairwise win probabilities, and
the resulting leaderboard obtained by sorting players according to
$\hat{\theta}$.

To keep the representation symmetric, every raw comparison is converted into a
two-row directed block. A decisive comparison in which player $i$ beats player
$j$ is represented as
\[
    i \succ j
    \quad \mapsto \quad
    B_b=\{(x_{ij},1), (x_{ji},0)\}.
\]
A tied comparison is represented as
\[
    i \sim j
    \quad \mapsto \quad
    B_b=\{(x_{ij},1), (x_{ji},1)\}.
\]
Thus, decisive outcomes provide one win and one loss across the two directions,
whereas ties assign equal win evidence to both players. This avoids choosing an
arbitrary winner for tied comparisons while keeping all raw comparisons in the
same binary directed BT input format.

For influence scoring and perturbation selection, we treat each two-row block
$B_b$ as the atomic unit. If the two directed rows corresponding to raw
comparison $b$ are indexed by $n_{ij}$ and $n_{ji}$, its block-level influence is
computed by summing the row-level influences,
\[
    I_{B_b}^{(f)}
    =
    I_{n_{ij}}^{(f)} + I_{n_{ji}}^{(f)} .
\]
All perturbation actions are then applied at the block level. A \Drop{} action
removes both directed rows in $B_b$; an \Add{} action inserts both directed rows
for the candidate comparison; and a \Flip{} action modifies the two directed
outcomes consistently. This ensures that influence scores and perturbation
budgets are defined at the level of the original raw comparisons, rather than at
the level of individual directed rows.

\subsection{One-step Newton refinement}
\label{app:1sn}

The influence approximation in Eq.~\ref{eq:propagate} linearizes the effect of
an infinitesimal change in a match weight. However, the actions used in our
experiments are finite. Following the classical one-step
case-deletion approximation for logistic regression diagnostics
\citep{pregibon_logistic_1981} and its use in influence-based leaderboard
analysis~\citep{huang_dropping_2025}, we use a one-step Newton (1sN)
correction to better approximate these finite actions without fully refitting
the BT model for every candidate.

For a comparison $z_n=(x_n,y_n)$ evaluated at the full-data fit $\hat{\theta}$,
let
\[
    p_n = \sigma(x_n^\top \hat{\theta}),
    \qquad
    r_n = y_n-p_n,
    \qquad
    v_n = p_n(1-p_n).
\]
Since
\[
    g_n
    =
    \nabla_\theta \ell(z_n;\hat{\theta})
    =
    (p_n-y_n)x_n
    =
    -r_nx_n,
\]
the first-order deletion approximation from Eq.~\ref{eq:param_influence} gives
\[
    \Delta\hat{\theta}^{\Drop}_{n,\mathrm{IF}}
    =
    -r_n H^{-1}x_n .
\]
For adding a candidate comparison $z_c=(x_c,y_c)$, the sign is reversed:
\[
    \Delta\hat{\theta}^{\Add}_{c,\mathrm{IF}}
    =
    r_c H^{-1}x_c .
\]

The influence approximation uses the original Hessian $H$ and therefore ignores the
fact that a finite \Drop or \Add action also changes the local curvature of the
BT objective. For a single comparison, this curvature change is rank-one because
the Hessian contribution of $z_n$ is
\[
    H_n = v_n x_nx_n^\top .
\]
Thus, dropping one comparison changes the local Hessian from $H$ to
$H-H_n$, while adding one comparison changes it from $H$ to $H+H_n$. Applying
the Sherman--Morrison \citep{sherman_morrison_1950} identity to this rank-one Hessian update yields the
leverage term
\[
    h_n = v_n x_n^\top H^{-1}x_n .
\]
This quantity measures how strongly comparison $n$ affects the local curvature
around the fitted BT solution.

The 1sN refinement rescales the influence update by this leverage term. For a single
dropped comparison,
\[
    \Delta\hat{\theta}^{\Drop}_{n,\mathrm{1sN}}
    =
    \frac{
        \Delta\hat{\theta}^{\Drop}_{n,\mathrm{IF}}
    }{
        1-h_n
    },
\]
whereas for a single added candidate comparison,
\[
    \Delta\hat{\theta}^{\Add}_{c,\mathrm{1sN}}
    =
    \frac{
        \Delta\hat{\theta}^{\Add}_{c,\mathrm{IF}}
    }{
        1+h_c
    }.
\]
Equivalently, 1sN performs one Newton step toward the optimum of the perturbed
objective, starting from the original fitted parameters $\hat{\theta}$. Compared
with influence, which keeps the curvature fixed at $H$, 1sN partially accounts for the
curvature change induced by the finite action while still avoiding a full BT
refit.

Given either the IF or 1sN parameter change, the predicted objective change is
computed using the same projection rule as Eq.~\ref{eq:propagate}:
\[
    \Delta f
    \approx
    \nabla_\theta f(\hat{\theta},w)^\top \Delta\hat{\theta}
    +
    \left(\frac{\partial f}{\partial w}\right)_{\mathrm{explicit}} .
\]
The explicit term is zero for objectives that depend on the data only through
$\hat{\theta}$, and nonzero for uncertainty objectives that directly depend on
the weighted comparison graph.

For Flip actions, the local Hessian is unchanged because it depends only on the
comparison features $x_n$ and not on the outcome $y_n$. As a result, the 1sN
correction does not apply, and the flip update reduces to its first-order
influence approximation.

\subsection{Grouped Newton refinement for player removal}
\label{app:group-newton}

The 1sN correction in Appendix~\ref{app:derivations} is applied to individual
\Drop or \Add actions. Player removal is a larger structured perturbation: for a
player $m$, it removes the full set of incident comparisons
\[
    \mathcal{G}_m
    =
    \{n : z_n \text{ contains player } m\}.
\]
Because many comparisons are removed simultaneously, we use a grouped Newton
refinement rather than applying the single-row leverage correction independently
to each match.

Let
\[
    p_n = \sigma(x_n^\top \hat{\theta}),
    \qquad
    r_n = y_n-p_n,
    \qquad
    s_n = r_n x_n
\]
denote the fitted probability, residual, and score contribution of comparison
$n$ at the full-data solution. For a group deletion $\mathcal{G}$, the first
step sums the removed score contributions and applies the grouped first-order
deletion update:
\[
    s_{\mathcal{G}}
    =
    \sum_{n\in\mathcal{G}} s_n,
    \qquad
    \theta_{\mathcal{G}}^{(1)}
    =
    \hat{\theta}
    -
    H^{-1}s_{\mathcal{G}} .
\]
This is the group analogue of summing individual deletion influences in
Eq.~\ref{eq:group_influence}.

The second step takes one Newton correction using only the comparisons that
remain after the group is removed. Let
\[
    \mathcal{K}
    =
    \{1,\dots,N\}\setminus \mathcal{G}
\]
be the kept set. At $\theta_{\mathcal{G}}^{(1)}$, define the kept-data score
\[
    S_{\mathcal{K}}(\theta_{\mathcal{G}}^{(1)})
    =
    \sum_{n\in\mathcal{K}}
    x_n\!\left(
    y_n-\sigma(x_n^\top \theta_{\mathcal{G}}^{(1)})
    \right),
\]
and the kept-data Hessian
\[
    H_{\mathcal{K}}(\theta_{\mathcal{G}}^{(1)})
    =
    \sum_{n\in\mathcal{K}}
    v_n^{(1)} x_nx_n^\top
    +
    \lambda I,
    \qquad
    v_n^{(1)}
    =
    p_n^{(1)}(1-p_n^{(1)}),
\]
where
\[
    p_n^{(1)}
    =
    \sigma(x_n^\top \theta_{\mathcal{G}}^{(1)}).
\]
The grouped Newton estimate is then
\[
    \theta_{\mathcal{G}}^{(2)}
    =
    \theta_{\mathcal{G}}^{(1)}
    +
    H_{\mathcal{K}}(\theta_{\mathcal{G}}^{(1)})^{-1}
    S_{\mathcal{K}}(\theta_{\mathcal{G}}^{(1)}).
\]

Finally, the predicted player-removal effect is evaluated by applying the
downstream objective to this approximate parameter vector:
\[
    \Delta f_{\mathcal{G}}
    \approx
    f(\theta_{\mathcal{G}}^{(2)}) - f(\hat{\theta}).
\]
For the player-removal experiments, $f$ is the smooth \KTau objective
computed over the remaining players, so the removed player is excluded from the
reference ranking before evaluating the objective.

This refinement is related to 1sN because both start from an influence-based
finite-deletion step and then use local curvature information to improve the
approximation. The difference is that 1sN uses a scalar leverage correction for
a single comparison, whereas the grouped Newton refinement performs one joint
Newton correction on the score equation of the remaining comparison data.

\subsection{Objective influence derivations}
\label{app:derivations}

The objective-level influence of a match $z_n$ is given by
Eq.~\ref{eq:propagate} and is defined as the derivative of the
objective with respect to increasing the match weight $w_n$. Thus,
for a finite \Drop{} action, where $\Delta w_n=-1$, the predicted
drop effect is
\[
\Delta f_{\mathrm{drop},n}
\approx
-\mathcal{I}_n^{(f)}.
\]
Equivalently, specialising to first-order case-deletion, the parameter
shift induced by dropping match \(n\) is
\[
\Delta\theta_n^{\mathrm{drop}} \approx -\,r_n H^{-1}x_n,
\qquad
r_n = y_n - p_n,
\qquad
p_n = \sigma(x_n^\top \hat\theta),
\]
and the corresponding drop effect is
\[
\Delta f_{\mathrm{drop},n}^{(f)}
\approx
\nabla_{\theta} f(\hat\theta,w)^\top
\Delta\theta_n^{\mathrm{drop}}
-
\left(\frac{\partial f}{\partial w_n}\right)_{\mathrm{explicit}}.
\]
Therefore,
\[
\mathcal{I}_n^{(f)}
\approx
-\Delta f_{\mathrm{drop},n}^{(f)}.
\]
For objectives that depend on the data only through \(\hat\theta\), the
explicit term is zero.

\paragraph{Top-$k$ membership objective / gap objective.}
For
\[
f_{\mathrm{gap}}(\theta)=\theta_i-\theta_j,
\]
the full gradient is
\[
\nabla_\theta f_{\mathrm{gap}} = e_i - e_j,
\]
and there is no explicit weight term, so the drop effect is
\[
\Delta f_{\mathrm{drop},n}^{(f_{\mathrm{gap}})}
=
\nabla_{\theta} f_{\mathrm{gap}}(\hat\theta)^\top
\Delta\theta_n^{\mathrm{drop}}.
\]
Equivalently,
\[
\mathcal{I}_n^{(f_{\mathrm{gap}})}
=
-\nabla_{\theta} f_{\mathrm{gap}}(\hat\theta)^\top
\Delta\theta_n^{\mathrm{drop}}.
\]

\paragraph{Player-uncertainty objective.}
The single-player uncertainty proxy is
\[
f_{\mathrm{CI\text{-}player}}(m;w)=\frac{1}{\hat{\rho}_m^2(w)},
\qquad
\hat{\rho}_m^2(w)=\sum_{j\neq m} w_{mj}\,v_{mj},
\]
with
\[
v_{mj}
=
\sigma(\hat\theta_m-\hat\theta_j)
\bigl(1-\sigma(\hat\theta_m-\hat\theta_j)\bigr).
\]
Define
\[
v'_{mj}
=
v_{mj}\bigl(1-2\sigma(\hat\theta_m-\hat\theta_j)\bigr).
\]
Then
\[
\frac{\partial \hat{\rho}_m^2}{\partial \theta_k}
=
\sum_{j\neq m} w_{mj}\,v'_{mj}\,
(\mathbf{1}_{k=m}-\mathbf{1}_{k=j}),
\]
so
\[
\nabla_\theta f_{\mathrm{CI\text{-}player}}(m;w)
=
-\frac{1}{(\hat{\rho}_m^2)^2}
\nabla_\theta \hat{\rho}_m^2.
\]
Unlike the gap and \KTau{} objectives, this objective has a nonzero
explicit weight derivative. If match \(n\) compares players \(a_n\)
and \(b_n\), then
\[
\left(\frac{\partial f_{\mathrm{CI\text{-}player}}(m;w)}
{\partial w_n}\right)_{\mathrm{explicit}}
=
-\frac{v_n\,\mathbf{1}\{m\in\{a_n,b_n\}\}}
{(\hat{\rho}_m^2)^2},
\]
where
\[
v_n=
\sigma(\hat\theta_{a_n}-\hat\theta_{b_n})
\bigl(1-\sigma(\hat\theta_{a_n}-\hat\theta_{b_n})\bigr).
\]
Therefore, the drop effect is
\[
\Delta f_{\mathrm{drop},n}^{(f_{\mathrm{CI\text{-}player}})}
=
\nabla_{\theta} f_{\mathrm{CI\text{-}player}}(m;w)^\top
\Delta\theta_n^{\mathrm{drop}}
+
\frac{v_n\,\mathbf{1}\{m\in\{a_n,b_n\}\}}
{(\hat{\rho}_m^2)^2}.
\]
Equivalently,
\[
\mathcal{I}_n^{(f_{\mathrm{CI\text{-}player}})}
=
-\nabla_{\theta} f_{\mathrm{CI\text{-}player}}(m;w)^\top
\Delta\theta_n^{\mathrm{drop}}
-
\frac{v_n\,\mathbf{1}\{m\in\{a_n,b_n\}\}}
{(\hat{\rho}_m^2)^2}.
\]

\paragraph{Trace-uncertainty objective.}
The global uncertainty proxy is
\[
f_{\mathrm{CI\text{-}trace}}(w)
=
\sum_{i=1}^M \frac{1}{\hat{\rho}_i^2(w)}.
\]
Hence
\[
\nabla_\theta f_{\mathrm{CI\text{-}trace}}
=
\sum_{i=1}^M
-\frac{1}{(\hat{\rho}_i^2)^2}
\nabla_\theta \hat{\rho}_i^2.
\]
For match \(n\) between players \(a_n\) and \(b_n\), the explicit
weight derivative is
\[
\left(\frac{\partial f_{\mathrm{CI\text{-}trace}}}
{\partial w_n}\right)_{\mathrm{explicit}}
=
-v_n
\left(
\frac{1}{(\hat{\rho}_{a_n}^2)^2}
+
\frac{1}{(\hat{\rho}_{b_n}^2)^2}
\right).
\]
Therefore, the drop effect is
\[
\Delta f_{\mathrm{drop},n}^{(f_{\mathrm{CI\text{-}trace}})}
=
\nabla_{\theta} f_{\mathrm{CI\text{-}trace}}(\hat\theta,w)^\top
\Delta\theta_n^{\mathrm{drop}}
+
v_n
\left(
\frac{1}{(\hat{\rho}_{a_n}^2)^2}
+
\frac{1}{(\hat{\rho}_{b_n}^2)^2}
\right).
\]
Equivalently,
\[
\mathcal{I}_n^{(f_{\mathrm{CI\text{-}trace}})}
=
-\nabla_{\theta} f_{\mathrm{CI\text{-}trace}}(\hat\theta,w)^\top
\Delta\theta_n^{\mathrm{drop}}
-
v_n
\left(
\frac{1}{(\hat{\rho}_{a_n}^2)^2}
+
\frac{1}{(\hat{\rho}_{b_n}^2)^2}
\right).
\]

\paragraph{\KTau{} surrogate.}
\[
f_{\tau,T}(\theta;\pi)
=
\frac{2}{M(M-1)}
\sum_{a<b}
s_{ab}\tanh\!\left(\frac{\theta_a-\theta_b}{T}\right),
\]
where \(s_{ab}\in\{-1,+1\}\) is induced by the reference ranking. Its
gradient is
\[
[\nabla_\theta f_{\tau,T}]_k
=
\frac{2}{M(M-1)}
\sum_{a<b}
s_{ab}
\frac{1-\tanh^2\!\left(\frac{\theta_a-\theta_b}{T}\right)}{T}
(\mathbf{1}_{k=a}-\mathbf{1}_{k=b}).
\]
There is no explicit weight term, so the drop effect is
\[
\Delta f_{\mathrm{drop},n}^{(f_{\tau,T})}
=
\nabla_{\theta} f_{\tau,T}(\hat\theta)^\top
\Delta\theta_n^{\mathrm{drop}}.
\]
Equivalently,
\[
\mathcal{I}_n^{(f_{\tau,T})}
=
-\nabla_{\theta} f_{\tau,T}(\hat\theta)^\top
\Delta\theta_n^{\mathrm{drop}}.
\]

\renewcommand{\algorithmicrequire}{\textbf{Input:}}
\renewcommand{\algorithmicensure}{\textbf{Output:}}

\subsection{Algorithms}
\label{app:algorithms}

\begin{algorithm}[H]
\caption{Top-$k$ Robustness Action Search}
\label{alg:topk-action-search}
\footnotesize
\begin{algorithmic}[1]
\REQUIRE Fitted BT model $\hat M$ on dataset $D$; top-$k$ cross-boundary candidate pairs $\mathcal P$; fixed action variant $a \in \{\texttt{drop}, \texttt{flip}, \texttt{add\_pairs}, \texttt{add\_outcomes}, \texttt{add\_weighted}\}$; maximum action count $A$
\ENSURE Smallest action count that makes at least one candidate top-$k$ boundary gap cross zero, together with the corresponding pair and selected actions
\STATE Fit the Bradley--Terry model on $D$ and obtain $\hat{\theta}$.
\STATE Construct the ordered set $\mathcal P$ of top-$k$ cross-boundary candidate pairs.
\STATE Initialize an empty cache of per-pair influence reports.
\FOR{$\alpha = 1,\dots,A$}
    \FORALL{$(i,j)\in\mathcal P$}
        \STATE Define the gap objective
        \[
            f_{ij}(\hat{\theta})=\hat{\theta}_i-\hat{\theta}_j .
        \]
        \STATE If not already cached, compute the one-step influence report for $(i,j)$ under action variant $a$ over its corresponding candidate pool.
        \STATE Let $g_{ij}=f_{ij}(\hat{\theta})$.
        \STATE Select the top $\alpha$ candidate actions from the cached report: if $g_{ij}>0$, choose the most negative influences; otherwise choose the most positive influences.
        \STATE Treat grouped forward/reverse copies as one logical action when applicable.
        \STATE Apply the selected $\alpha$ actions to the original fitted dataset $D$, refit the BT model, and compute the updated gap $g_{ij}^{(\alpha)}$ or estimate it using influence scores.
        \IF{$(g_{ij}>0 \ \mathrm{and}\ g_{ij}^{(\alpha)} \le 0)$ or $(g_{ij}<0 \ \mathrm{and}\ g_{ij}^{(\alpha)} \ge 0)$}
            \STATE \textbf{return} $\alpha$, the pair $(i,j)$, and the selected actions.
        \ENDIF
    \ENDFOR
\ENDFOR
\STATE \textbf{return} failure if no pair succeeds within $A$ actions.
\end{algorithmic}
\end{algorithm}


\begin{algorithm}[H]
\caption{Strict CI-Aware Top-$k$ Manipulation}
\label{alg:strict-ci-topk}
\footnotesize
\begin{algorithmic}[1]
\REQUIRE Fitted BT model $\hat M$ on dataset $D$; target boundary rank $k$; fixed insider--outsider pair $(i,j)$ with $i$ the rank-$k$ model and $j$ the rank-$(k+1)$ model; CI method; CI level $\alpha_{\mathrm{CI}}$; fixed action variant $a \in \{\texttt{drop}, \texttt{flip}, \texttt{add\_pairs}, \texttt{add\_outcomes}, \texttt{add\_weighted}\}$; maximum action budget $A$
\ENSURE Minimum number of actions needed to certify that outsider $j$ lies above insider $i$ under strict CI separation
\STATE Fit the Bradley--Terry model on $D$ and compute $\hat{\theta}$ and confidence intervals
\[
  [L_m,U_m]
  =
  \left[
    \hat{\theta}_m - z_{\alpha_{\mathrm{CI}}}\mathrm{SE}_m,\;
    \hat{\theta}_m + z_{\alpha_{\mathrm{CI}}}\mathrm{SE}_m
  \right],
  \qquad
  z_{\alpha_{\mathrm{CI}}}=\Phi^{-1}(1-\alpha_{\mathrm{CI}}/2).
\]
\STATE Fix the target pair $(i,j)$ before the action search begins, where $i$ is the current insider and $j$ is the current outsider.
\STATE Define the strict CI objective
\[
  g_{ij}(\hat{\theta})
  =
  \bigl(\hat{\theta}_i + z_{\alpha_{\mathrm{CI}}}\mathrm{SE}_i\bigr)
  -
  \bigl(\hat{\theta}_j - z_{\alpha_{\mathrm{CI}}}\mathrm{SE}_j\bigr).
\]
\STATE Proceed only if the strict target is initially unmet: $g_{ij}(\hat{\theta}) \ge 0$.
\STATE Compute the full one-step influence report for action variant $a$ with respect to $g_{ij}$ over its corresponding candidate pool.
\FOR{$\ell=1,\dots,A$}
    \STATE Select the top $\ell$ logical actions that most decrease $g_{ij}$.
    \STATE Rank candidates in ascending order of influence and group forward/reverse copies sharing a match identifier as one logical action.
    \STATE Form the first-order screened objective
    \[
        \widetilde g_{ij}^{(\ell)}
        =
        g_{ij}(\hat\theta)
        +
        \sum_{n\in S_{ij}^{(\ell)}} \mathcal{I}^{(g_{ij})}_n .
    \]
    \IF{$\widetilde g_{ij}^{(\ell)} \ge 0$}
        \STATE Continue to the next budget without refitting.
    \ENDIF
    \STATE Apply the selected actions to the original fitted dataset and refit the BT model.
    \STATE Recompute the exact strict objective on the refit model:
    \[
      g_{ij}^{\mathrm{refit}}
      =
      \bigl(\hat{\theta}_i^{\,\mathrm{refit}}
      + z_{\alpha_{\mathrm{CI}}}\mathrm{SE}_i^{\,\mathrm{refit}}\bigr)
      -
      \bigl(\hat{\theta}_j^{\,\mathrm{refit}}
      - z_{\alpha_{\mathrm{CI}}}\mathrm{SE}_j^{\,\mathrm{refit}}\bigr).
    \]
    \STATE Let
    \[
      L_j^{\mathrm{refit}}
      =
      \hat{\theta}_j^{\,\mathrm{refit}}
      -
      z_{\alpha_{\mathrm{CI}}}\mathrm{SE}_j^{\,\mathrm{refit}},
      \qquad
      U_i^{\mathrm{refit}}
      =
      \hat{\theta}_i^{\,\mathrm{refit}}
      +
      z_{\alpha_{\mathrm{CI}}}\mathrm{SE}_i^{\,\mathrm{refit}}.
    \]
    \IF{$g_{ij}^{\mathrm{refit}} < 0$, equivalently $L_j^{\mathrm{refit}} > U_i^{\mathrm{refit}}$}
        \STATE \textbf{return} $\ell$, the pair $(i,j)$, the selected actions, and the refit ranking.
    \ENDIF
\ENDFOR
\STATE \textbf{return} failure if no strict CI-certified reversal is found within budget $A$.
\end{algorithmic}
\end{algorithm}


\begin{algorithm}[H]
\caption{Strict CI-Aware $k$-Selection}
\label{alg:strict-ci-k-selection}
\footnotesize
\begin{algorithmic}[1]
\REQUIRE Fitted BT model $\hat M$ on dataset $D$; CI method; CI level $\alpha_{\mathrm{CI}}$
\ENSURE One valid target triple $(k,i,j)$ for strict CI-aware manipulation, where $i$ is the insider at rank $k$ and $j$ is the outsider at rank $k+1$
\STATE Fit the Bradley--Terry model on $D$ and compute the CI ranking.
\FOR{$k=1,\dots,N-1$}
    \STATE Let
    \[
      i=\text{rank-}k\text{ model},
      \qquad
      j=\text{rank-}(k+1)\text{ model}.
    \]
    \STATE Define the strict CI objective
    \[
      g_{ij}(\hat{\theta})
      =
      \bigl(\hat{\theta}_i + z_{\alpha_{\mathrm{CI}}}\mathrm{SE}_i\bigr)
      -
      \bigl(\hat{\theta}_j - z_{\alpha_{\mathrm{CI}}}\mathrm{SE}_j\bigr),
      \qquad
      z_{\alpha_{\mathrm{CI}}}=\Phi^{-1}(1-\alpha_{\mathrm{CI}}/2).
    \]
    \STATE Keep boundary $(k,i,j)$ only if $g_{ij}(\hat{\theta}) \ge 0$.
\ENDFOR
\STATE \textbf{return} the valid boundary $(k,i,j)$ with the smallest strict objective value $g_{ij}(\hat{\theta})$.
\STATE \textbf{return} failure if no valid boundary satisfies $g_{ij}(\hat{\theta}) \ge 0$.
\end{algorithmic}
\end{algorithm}


\begin{algorithm}[H]
\caption{Online Rigging Baseline (Omni-On)}
\label{alg:online-rigging-omni-on}
\footnotesize
\begin{algorithmic}[1]
\REQUIRE Initial Elo/BT ranking on dataset $D$; target model $m^*$; direction $d\in\{\texttt{promote},\texttt{demote}\}$; exposed pair stream $\{(a_t,b_t)\}_{t=1}^B$; budget $B$; Elo constants $K,\mathrm{BASE},\mathrm{SCALE}$
\ENSURE Ordered decisions and target-rank trajectory
\STATE Compute the initial ratings $\hat{\theta}_m$ for all models and record the initial rank of $m^*$.
\FOR{$t=1,\dots,B$}
    \STATE Observe the currently exposed pair $(a_t,b_t)$.
    \STATE Define the candidate decision set
    \[
        \mathcal{A}_t
        =
        \{\texttt{a\_wins},\ \texttt{b\_wins},\ \texttt{tie},\ \texttt{remove}\}.
    \]
    \STATE Let $r_a=\hat{\theta}_{a_t}$, $r_b=\hat{\theta}_{b_t}$, and $r_*=\hat{\theta}_{m^*}$, and compute
    \[
        e_a=\frac{1}{1+\mathrm{BASE}^{(r_b-r_a)/\mathrm{SCALE}}},
        \qquad
        e_b=\frac{1}{1+\mathrm{BASE}^{(r_a-r_b)/\mathrm{SCALE}}}.
    \]
    \FORALL{$\alpha \in \mathcal{A}_t$}
        \STATE Form one-step hypothetical ratings:
        \[
        \begin{array}{ll}
        \texttt{a\_wins}: &
        r_a' = r_a + K e_b,\quad r_b' = r_b - K e_b, \\[2pt]
        \texttt{b\_wins}: &
        r_a' = r_a - K e_a,\quad r_b' = r_b + K e_a, \\[2pt]
        \texttt{tie}: &
        r_a' = r_a - \frac{K}{2}(e_a-e_b),\quad
        r_b' = r_b + \frac{K}{2}(e_a-e_b), \\[2pt]
        \texttt{remove}: &
        r_a' = r_a,\quad r_b' = r_b .
        \end{array}
        \]
        \STATE Score the action by
        \[
        r_t^{(\alpha)}
        =
        \frac{1}{1+\mathrm{BASE}^{(r_a'-r_*)/\mathrm{SCALE}}}
        +
        \frac{1}{1+\mathrm{BASE}^{(r_b'-r_*)/\mathrm{SCALE}}}.
        \]
        \STATE For demotion tasks, use the inverted reward $-r_t^{(\alpha)}$.
    \ENDFOR
    \STATE Select the greedy rigging decision
    \[
        \alpha_t^*=\arg\max_{\alpha \in \mathcal{A}_t} r_t^{(\alpha)}.
    \]
    \STATE Apply $\alpha_t^*$ to the current dataset, refit the global Elo/BT ranking, and append the new rank of $m^*$ to the history.
\ENDFOR
\STATE \textbf{return} the decision history and the full target-rank trajectory.
\end{algorithmic}
\end{algorithm}


\begin{algorithm}[H]
\caption{Online Influence-Guided Targeted Top-$k$ Manipulation}
\label{alg:online-influence-targeted-topk}
\footnotesize
\begin{algorithmic}[1]
\REQUIRE Fitted BT model on dataset $D$; target model $m^*$; rank cutoff $k$; direction $d\in\{\texttt{promote},\texttt{demote}\}$; exposed pair stream $\{(a_t,b_t)\}_{t=1}^B$; budget $B$
\ENSURE Selected decision sequence and final ranking
\STATE Fit the initial BT model on $D$ and record the initial ranking of $m^*$.
\FOR{$t=1,\dots,B$}
    \IF{$\operatorname{rank}(m^*)\le k$ for promotion, or $\operatorname{rank}(m^*)>k$ for demotion}
        \STATE Stop; the target condition is already satisfied.
    \ENDIF
    \STATE Observe the currently exposed pair $(a_t,b_t)$.
    \STATE Define the current boundary objective
    \[
        f_t(\hat{\theta}) =
        \begin{cases}
        \hat{\theta}_{m^*}-\hat{\theta}_{m_k}, & d=\texttt{promote},\\[3pt]
        \hat{\theta}_{m_{k+1}}-\hat{\theta}_{m^*}, & d=\texttt{demote},
        \end{cases}
    \]
    where $m_k$ and $m_{k+1}$ are recomputed from the current ranking after every accepted intervention.
    \STATE Form the allowable decision set
    \[
        \mathcal{A}_t
        =
        \{\texttt{remove},\ \texttt{a\_wins},\ \texttt{b\_wins},\ \texttt{tie}\}.
    \]
    \FORALL{$\alpha\in\mathcal{A}_t$}
        \STATE Compute its first-order influence score on the current boundary objective:
        \[
        \mathcal{I}^{(f_t)}(\alpha)
        =
        \begin{cases}
        0, & \alpha=\texttt{remove},\\
        \mathcal{I}^{(f_t)}(a_t \succ b_t), & \alpha=\texttt{a\_wins},\\
        \mathcal{I}^{(f_t)}(b_t \succ a_t), & \alpha=\texttt{b\_wins},\\
        \mathcal{I}^{(f_t)}(a_t \succ b_t)
        +
        \mathcal{I}^{(f_t)}(b_t \succ a_t), & \alpha=\texttt{tie}.
        \end{cases}
        \]
    \ENDFOR
    \STATE Select the greedy influence decision
    \[
        \alpha_t^*=\arg\max_{\alpha\in\mathcal{A}_t} \mathcal{I}^{(f_t)}(\alpha).
    \]
    \STATE Apply $\alpha_t^*$ to the current state, refit the BT model if any row is added or estimate it using influence scores, and update the ranking of $m^*$.
\ENDFOR
\STATE \textbf{return} the selected decisions, success indicator, rank trajectory, and final ranking.
\end{algorithmic}
\end{algorithm}


\begin{algorithm}[H]
\caption{CI Reduction via Targeted Match Addition}
\label{alg:ci-reduction-addition}
\footnotesize
\begin{algorithmic}[1]
\REQUIRE Fitted dataset $D$; target model $m^*$; budget $B$; candidate mode $v$; CI method
\ENSURE Added matches and CI-width trajectory
\STATE Fit the BT model on $D$.
\STATE Define the target uncertainty objective
\[
    f_{\mathrm{CI\text{-}player}}(m^*;w)
    =
    \hat{\rho}_{m^*}^{-2}(w),
\]
where $\hat{\rho}_{m^*}^2(w)$ is the BT pair-weight-aggregated variance proxy defined in Eq.~\ref{eq:local_info_rho}.
\STATE Generate add candidates according to $v$.
\IF{$v=\texttt{all\_pairs}$}
    \STATE Use one unordered pair, with the currently higher-rated model as winner.
\ENDIF
\STATE Compute influence scores for all add candidates once using the initial fitted model.
\STATE Sort candidates in ascending influence order, since negative influence reduces $f_{\mathrm{CI\text{-}player}}$.
\FOR{$t=1,\dots,B$}
    \STATE Select the next unused candidate from the fixed sorted list.
    \STATE Add the match to $D$, refit the BT model or estimate it using influence scores, and recompute the target CI width using the chosen CI method.
\ENDFOR
\STATE \textbf{return} the added matches, uncertainty trajectory, and CI-width trajectory.
\end{algorithmic}
\end{algorithm}

\subsection{Relation between the sandwich covariance and Gao-style local information}
\label{app:sandwich_gao_relation}

In the main text, we use the Gao-style local information approximation as a
scalable uncertainty proxy for BT leaderboard perturbations. Here, we clarify how
this approximation relates to the more general sandwich covariance estimator and
why we avoid propagating perturbations through the full covariance matrix.

For a comparison $z_n=(i_n,j_n,y_n)$, define
$x_n=e_{i_n}-e_{j_n}$,
$p_n=\sigma(x_n^\top\hat{\theta})$, and
$v_n=p_n(1-p_n)$.
The weighted BT loss has score and Hessian
\begin{equation}
    g_n(\hat{\theta})=(p_n-y_n)x_n,
    \qquad
    H_n(\hat{\theta})=v_n x_nx_n^\top .
    \label{eq:bt_score_hessian_appendix}
\end{equation}
The sandwich covariance estimator is
\begin{equation}
    \Sigma_{\mathrm{sand}}(w)
    =
    J(w)^{-1}S(w)J(w)^{-1},
    \qquad
    J(w)=\sum_{n=1}^{N}w_nH_n(\hat{\theta})+\lambda I,
    \qquad
    S(w)=\sum_{n=1}^{N}w_ng_n(\hat{\theta})g_n(\hat{\theta})^\top .
    \label{eq:sandwich_appendix}
\end{equation}
Under a correctly specified BT model,
\[
    \mathbb{E}\!\left[g_n(\theta^\star)g_n(\theta^\star)^\top
    \mid i_n,j_n\right]
    =
    v_n x_nx_n^\top
    =
    H_n(\theta^\star),
\]
because
$\mathrm{Var}(y_n\mid i_n,j_n)=p_n(1-p_n)=v_n$.
Thus, replacing the empirical score covariance $S(w)$ by its information
counterpart gives $S(w)\approx J(w)$, so the sandwich estimator reduces to the information-based covariance
approximation
\begin{equation}
    \Sigma_{\mathrm{sand}}(w)
    =
    J(w)^{-1}S(w)J(w)^{-1}
    \approx
    J(w)^{-1}.
    \label{eq:sandwich_to_info_appendix}
\end{equation}
Gao et al.~\cite{gao_uncertainty_2022} provide a coordinate-wise interpretation
of uncertainty in the BT model: the leading uncertainty of each skill estimate is
controlled by the inverse of a local Fisher information term. In our weighted
finite-sample setting, this local information for player $i$ is
\begin{equation}
    \hat{\rho}_i^2(w)
    =
    \sum_{j\neq i} w_{ij}
    p_{ij}(1-p_{ij}),
    \label{eq:rho_weighted_appendix}
\end{equation}

Equivalently, if
\[
    J_{\mathrm{info}}(w)
    =
    \sum_{i<j}w_{ij}p_{ij}(1-p_{ij})
    (e_i-e_j)(e_i-e_j)^\top+\lambda I,
\]
then $\hat{\rho}_i^2(w)$ is the local diagonal information associated with
player $i$.

The full inverse $J_{\mathrm{info}}(w)^{-1}$ captures coupling across all
players, but computing and differentiating its diagonal or trace under many
candidate perturbations is substantially more expensive. We therefore use the
Gao-style local approximation
\[
    \mathrm{Var}(\hat{\theta}_i)
    \approx
    \hat{\rho}_i^{-2}(w),
\]
which leads directly to the player-wise uncertainty objective
$f_{\mathrm{CI\text{-}player}}(i;w)=\hat{\rho}_i^{-2}(w)$
and the trace-style global uncertainty proxy
$f_{\mathrm{CI\text{-}trace}}(w)=\sum_i\hat{\rho}_i^{-2}(w)$.
This approximation preserves the main statistical intuition: uncertainty is
reduced by adding informative comparisons incident to poorly measured players,
and increased by removing such comparisons.

\subsection{Arena Active baseline for target-model CI reduction}
\label{app:arena-active}

Arena Active~\citep{chiang_chatbot_2024} is implemented as a target-restricted 
variance-count active-sampling baseline. At each step, it recomputes scores 
under the current Bradley--Terry fit, considers only candidate pairs involving 
the target model $m^{*}$, and selects the opponent $j$ with the largest estimated 
one-step reduction
\begin{equation}
    s_{\mathrm{AA}}(m^{*}, j)
    =
    \sqrt{\frac{\widehat{\mathrm{Var}}(\hat{\theta}_{m^{*}}-\hat{\theta}_j)}
    {N_{m^{*}j}}}
    -
    \sqrt{\frac{\widehat{\mathrm{Var}}(\hat{\theta}_{m^{*}}-\hat{\theta}_j)}
    {N_{m^{*}j}+1}},
\end{equation}
where $N_{m^{*}j}$ is the current number of comparisons between $m^{*}$ and 
$j$. Thus, Arena Active is target-specific through candidate filtering, but 
uses a pair-level variance-count score rather than directly optimizing 
$f_{\mathrm{CI\text{-}player}}(m^{*})$. In contrast, our method scores each 
candidate \Add action by its predicted effect on the target uncertainty 
objective.

Arena Active always uses an \texttt{all\_pairs} candidate space, while the 
influence policy can use \texttt{all\_pairs}, \texttt{all\_outcomes}, or 
\texttt{all\_outcomes\_weighted}. Thus, \texttt{all\_pairs} is the fairest 
direct comparison, whereas the outcome-aware spaces evaluate a richer influence 
action space.
\section{Reproducibility Details}
\label{app:reproducibility}

For reproducibility, we report the main hyperparameters and compute settings used in our experiments. Unless otherwise stated, all Bradley--Terry models were fit with \texttt{hessian\_ridge}=0.0. In the grouped Newton refinement, we used a small numerical stabilization ridge of \(10^{-8}\). For the smooth Kendall's \(\tau\) surrogate, we used temperature \(T=0.1\) in the player-level grouped-drop analysis and \(T=0.5\) in the robustness and action-curve analyses. All influence computations used the \texttt{1sn} approximation throughout.


All main experiments were run with 4 CPUs and 32\,GB memory per task. We used the same hyperparameter settings across datasets unless explicitly noted above.

\section{Other Results}

\subsection{{Dataset-level robustness scores}}
\label{app:robustness-scores}

To summarize robustness across datasets, we define normalized audit scores for
the three leaderboard conclusions studied in the main text. Let $B$ denote the
maximum perturbation budget used for the corresponding audit and let $b^\star$
be the minimum number of influence-guided actions required to change the audited
criterion after refitting the Bradley--Terry model. If no change occurs within
the budget, we set $b^\star=B$. We define
\[
R_{{\mathrm{{Top}}\text{{-}}1}} = \frac{{b^\star_{{\mathrm{{Top}}\text{{-}}1}}}}{{B_{{\mathrm{{Top}}\text{{-}}1}}}},
\qquad
R_{{\mathrm{{CITrace}}}} = \frac{{U(B_{{\mathrm{{CITrace}}}})}}{{U(0)}},
\qquad
R_{{\tau}} = \tau(B_\tau),
\]
where $R_{{\mathrm{{Top}}\text{{-}}1}}$ measures the normalized budget needed to change
the point-estimate Top-1 boundary, $R_{{\mathrm{{CITrace}}}}$ reports the remaining
trace-uncertainty proxy under the fixed CI-trace audit budget, and $R_\tau$
reports the Kendall-$\tau$ value after the fixed global-ranking audit budget.
Thus, lower $R_{{\mathrm{{Top}}\text{{-}}1}}$ indicates that fewer actions are needed to
change the local leaderboard conclusion, while lower $R_{{\mathrm{{CITrace}}}}$ and
lower $R_\tau$ indicate stronger degradation under the corresponding fixed-budget
audits.

For a compact overall summary, we also report
\begin{equation}
R_{\mathrm{all}}
=
\frac{1}{3}
\left(
R_{\mathrm{Top}\text{-}1}
+
R_{\mathrm{CITrace}}
+
R_{\tau}
\right).
\label{eq:overall-robustness}
\end{equation}
This aggregate is not intended to replace the three component scores. Instead,
it provides a single coarse audit number while preserving the decomposition into
local point-estimate fragility, CI-trace robustness, and global ranking stability.


\subsection{\texorpdfstring{\KTau}{Kendall's tau} and trace uncertainty curves}
\label{app:tau-trace-curves}

\begin{figure}[H]
    \centering
    \begin{subfigure}{0.48\linewidth}
        \includegraphics[width=\linewidth,trim=0 0 0 8mm, clip]{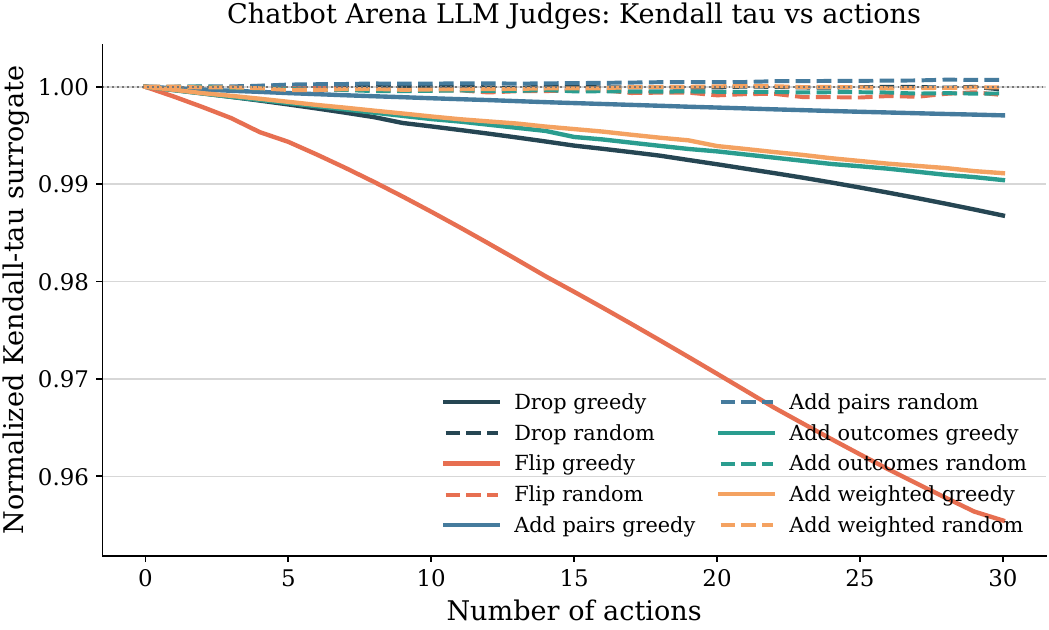}
        \caption{\KTau.}
    \end{subfigure}
    \hfill
    \begin{subfigure}{0.48\linewidth}
        \includegraphics[width=\linewidth,trim=0 0 0 8mm, clip]{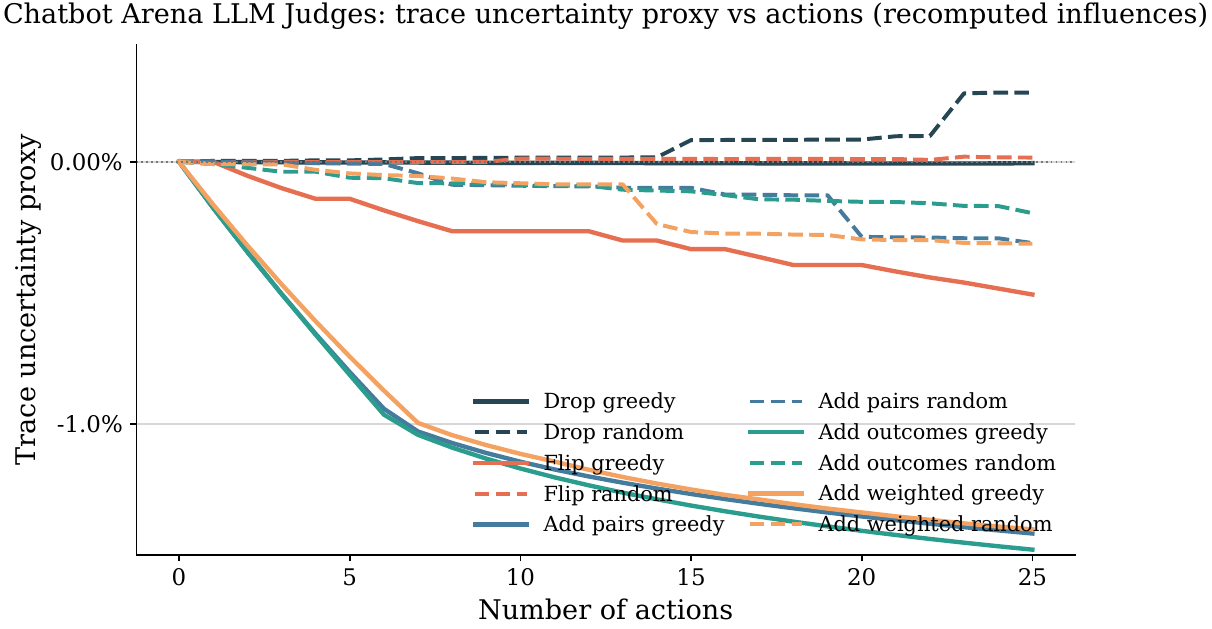}
        \caption{Trace uncertainty (\%).}
    \end{subfigure}
    \caption{Perturbation-budget curves for \textbf{Chatbot Arena LLM Judges} (49{,}938 matches, 64 models). Influence-guided Flip degrades $\tau$ most steeply; Add greedy variants dominate uncertainty reduction.}
    \label{fig:app_llm_judge}
\end{figure}

\begin{figure}[H]
    \centering
    \begin{subfigure}{0.48\linewidth}
        \includegraphics[width=\linewidth,trim=0 0 0 8mm, clip]{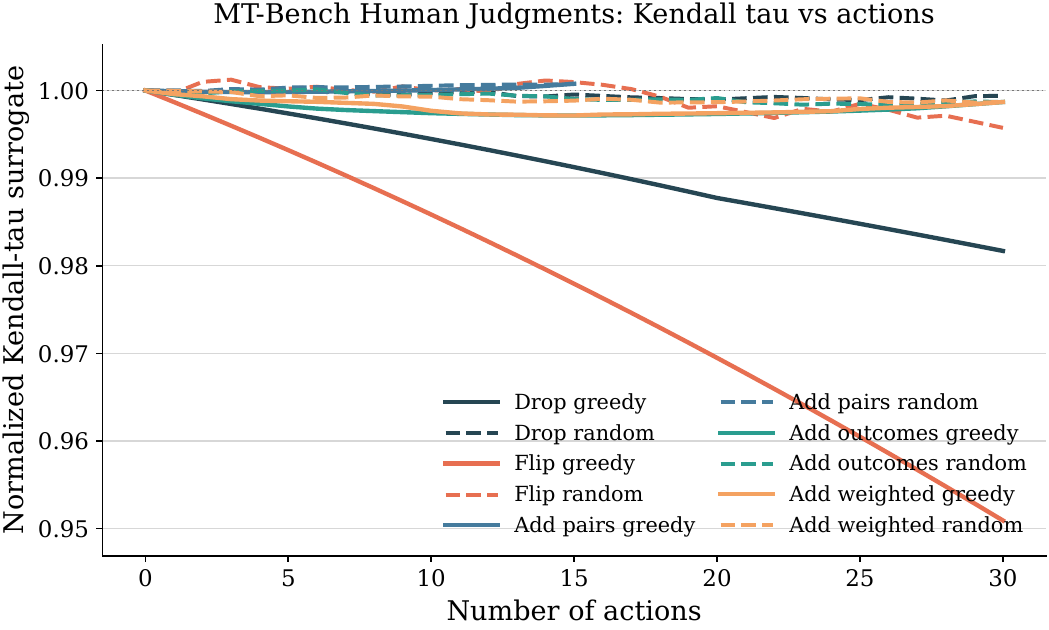}
        \caption{\KTau.}
    \end{subfigure}
    \hfill
    \begin{subfigure}{0.48\linewidth}
        \includegraphics[width=\linewidth,trim=0 0 0 8mm, clip]{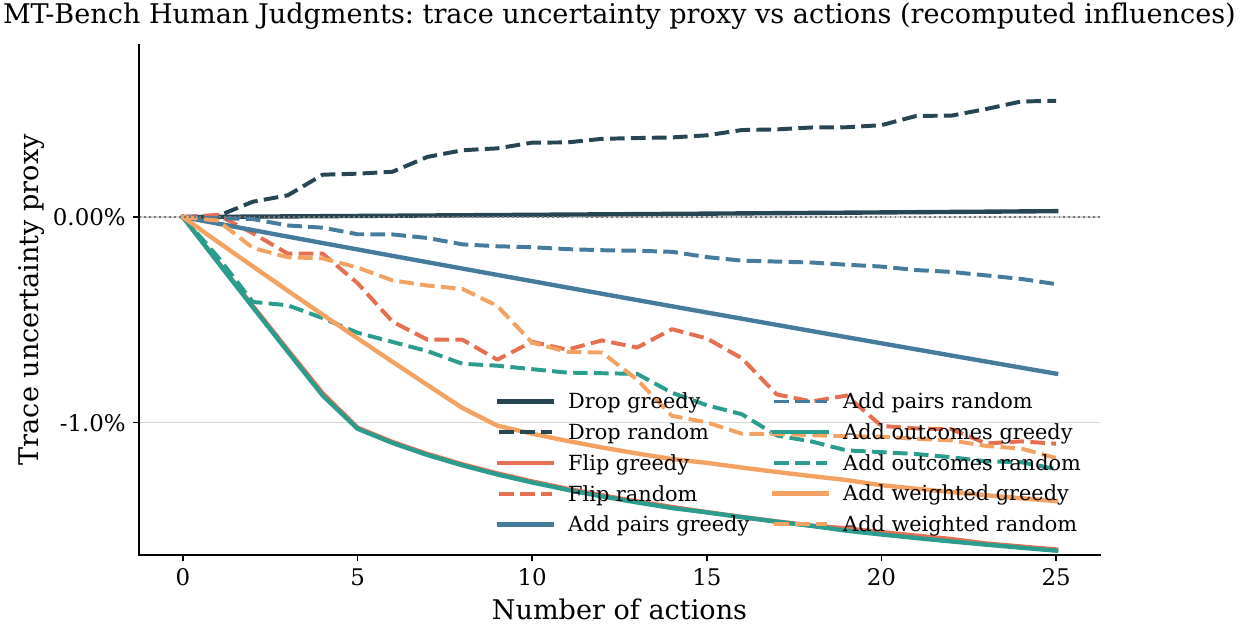}
        \caption{Trace uncertainty (\%).}
    \end{subfigure}
    \caption{Perturbation-budget curves for \textbf{MT-Bench Human Judgments} (3{,}355 matches). $\tau$ degrades more slowly than on crowd-sourced datasets, consistent with Table~\ref{tab:actions_summary} showing MT-Bench is the most robust dataset.}
    \label{fig:app_mt_bench}
\end{figure}

\begin{figure}[H]
    \centering
    \begin{subfigure}{0.48\linewidth}
        \includegraphics[width=\linewidth,trim=0 0 0 8mm, clip]{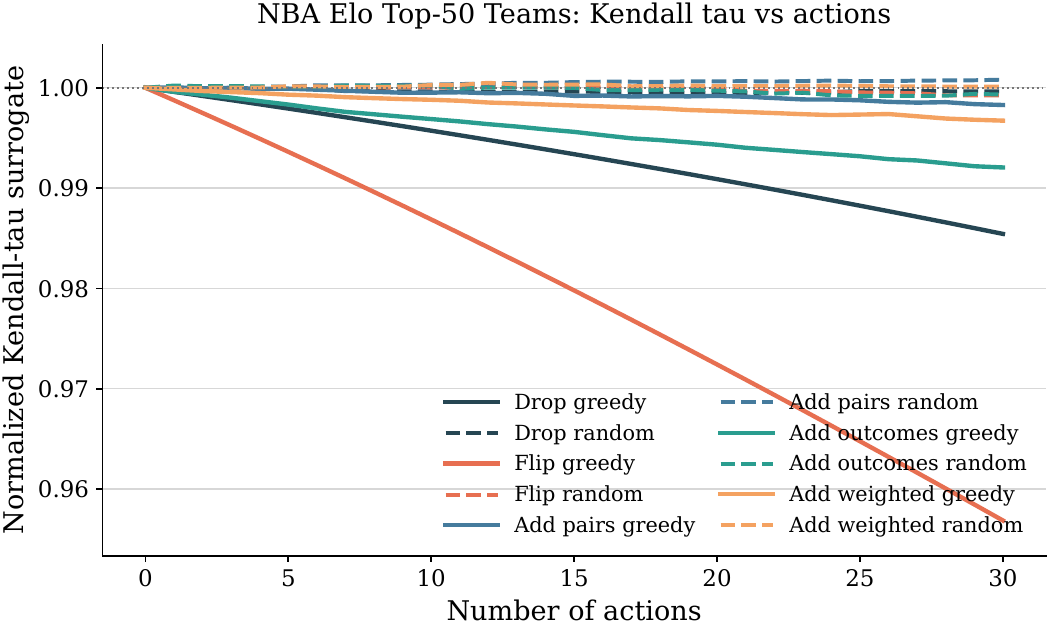}
        \caption{\KTau.}
    \end{subfigure}
    \hfill
    \begin{subfigure}{0.48\linewidth}
        \includegraphics[width=\linewidth,trim=0 0 0 8mm, clip]{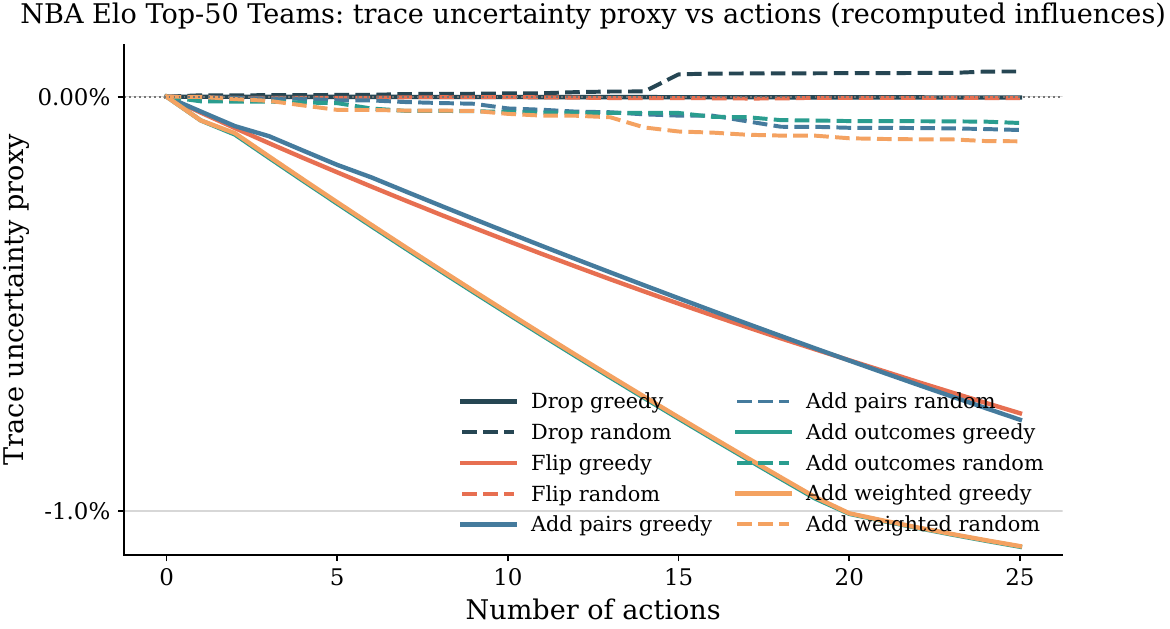}
        \caption{Trace uncertainty (\%).}
    \end{subfigure}
    \caption{Perturbation-budget curves for \textbf{NBA Elo Top-50 Teams} (109{,}892 matches). The dense graph limits per-action impact on $\tau$, but the Add-dominates-uncertainty reversal persists.}
    \label{fig:app_nba}
\end{figure}

\begin{figure}[H]
    \centering
    \begin{subfigure}{0.48\linewidth}
        \includegraphics[width=\linewidth,trim=0 0 0 8mm, clip]{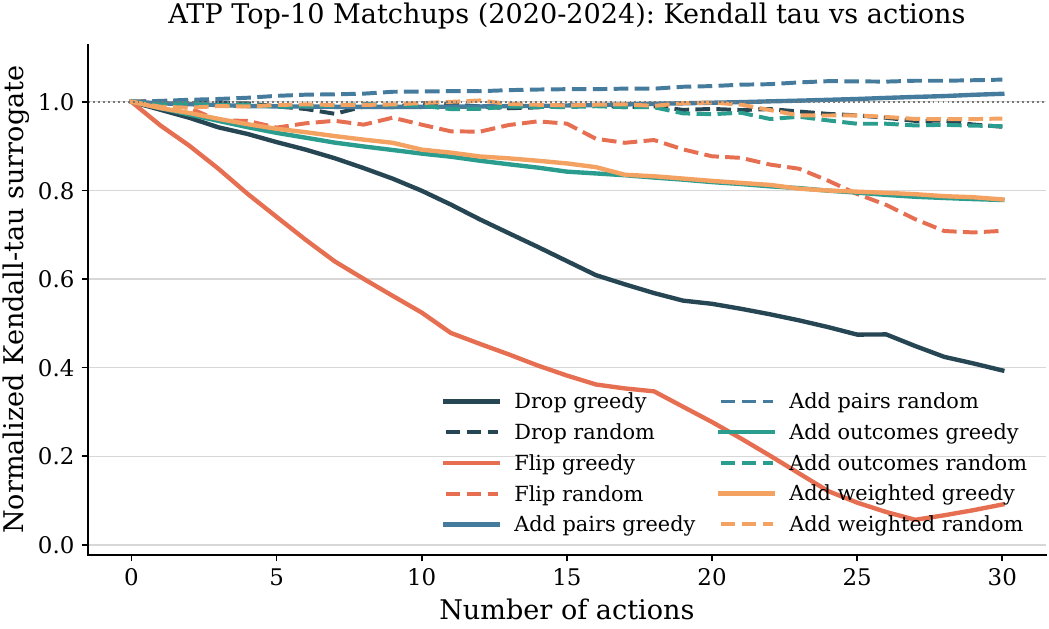}
        \caption{\KTau.}
    \end{subfigure}
    \hfill
    \begin{subfigure}{0.48\linewidth}
        \includegraphics[width=\linewidth,trim=0 0 0 8mm, clip]{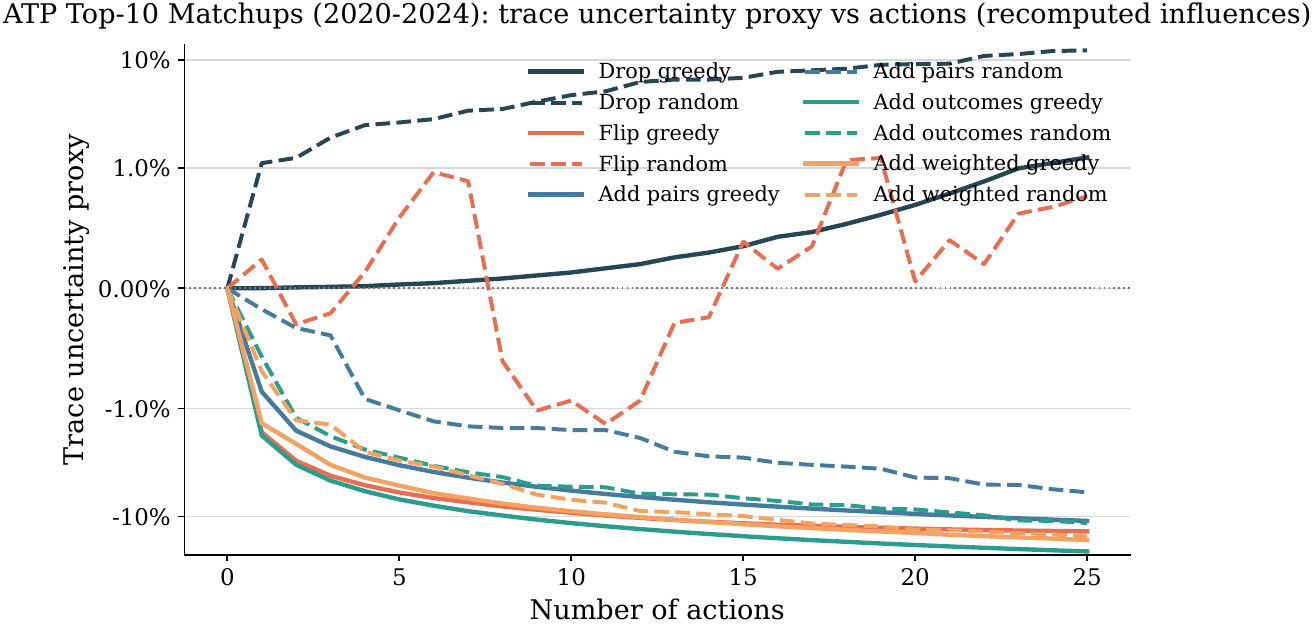}
        \caption{Trace uncertainty (\%).}
    \end{subfigure}
    \caption{Perturbation-budget curves for \textbf{ATP Top-10 Matchups} (278 matches). The extremely sparse graph makes this the least robust dataset: Flip drives $\tau$ to near 0 within 30 actions.}
    \label{fig:app_atp}
\end{figure}

\begin{figure}[H]
    \centering
    \begin{subfigure}{0.48\linewidth}
        \includegraphics[width=\linewidth,trim=0 0 0 8mm, clip]{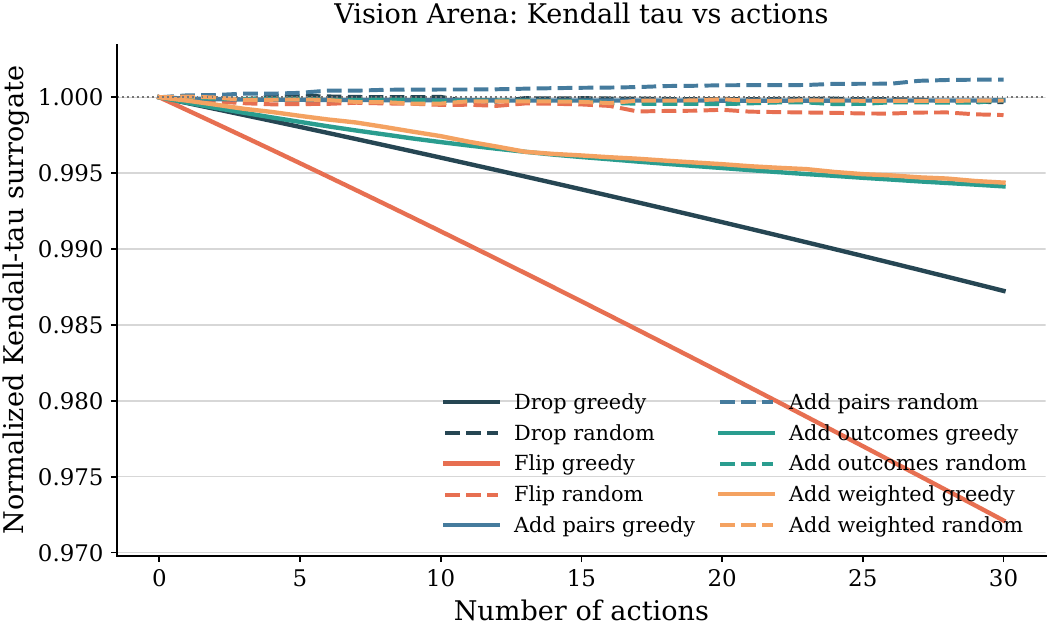}
        \caption{\KTau.}
    \end{subfigure}
    \hfill
    \begin{subfigure}{0.48\linewidth}
        \includegraphics[width=\linewidth,trim=0 0 0 8mm, clip]{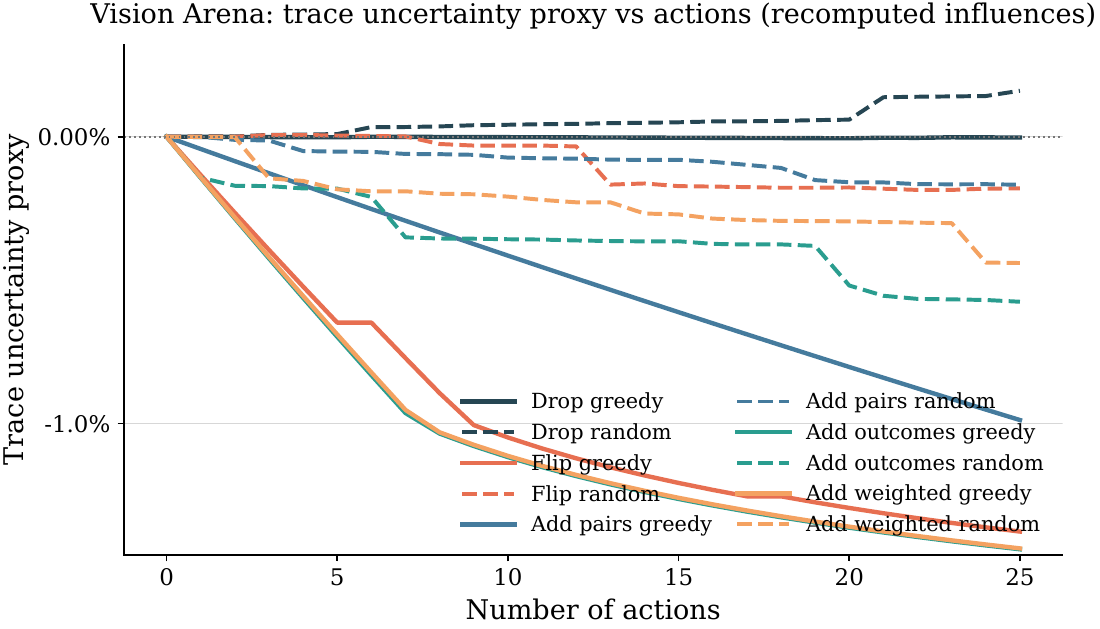}
        \caption{Trace uncertainty (\%).}
    \end{subfigure}
    \caption{Perturbation-budget curves for \textbf{Vision Arena} (29{,}849 matches). Patterns mirror those of Arena 55k: Flip leads $\tau$ degradation and Add variants dominate uncertainty.}
    \label{fig:app_vision}
\end{figure}

\begin{figure}[H]
    \centering
    \begin{subfigure}{0.48\linewidth}
        \includegraphics[width=\linewidth,trim=0 0 0 8mm, clip]{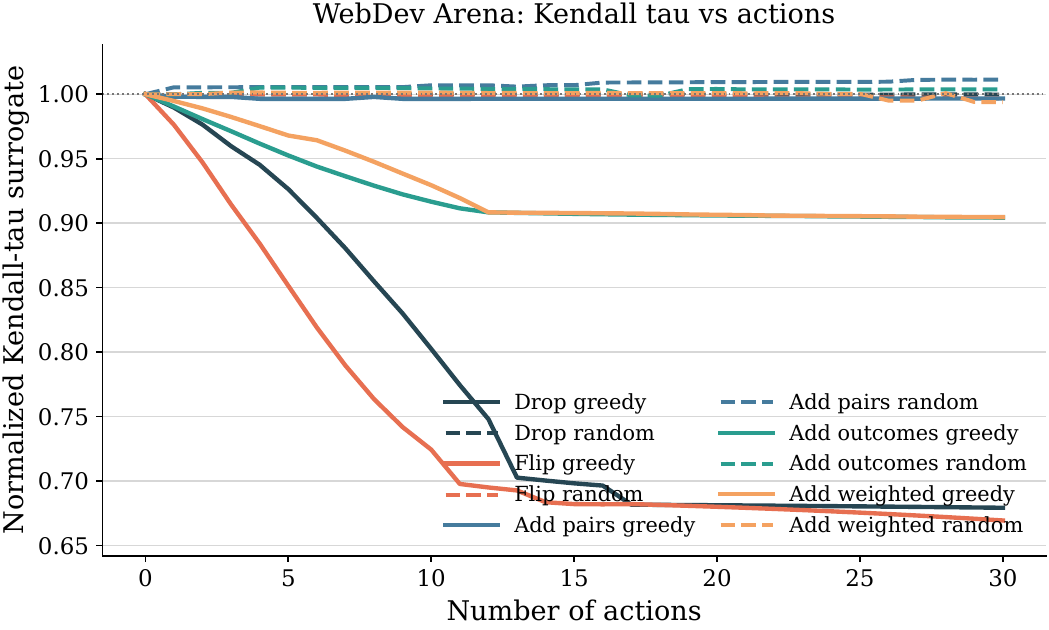}
        \caption{\KTau.}
    \end{subfigure}
    \hfill
    \begin{subfigure}{0.48\linewidth}
        \includegraphics[width=\linewidth,trim=0 0 0 8mm, clip]{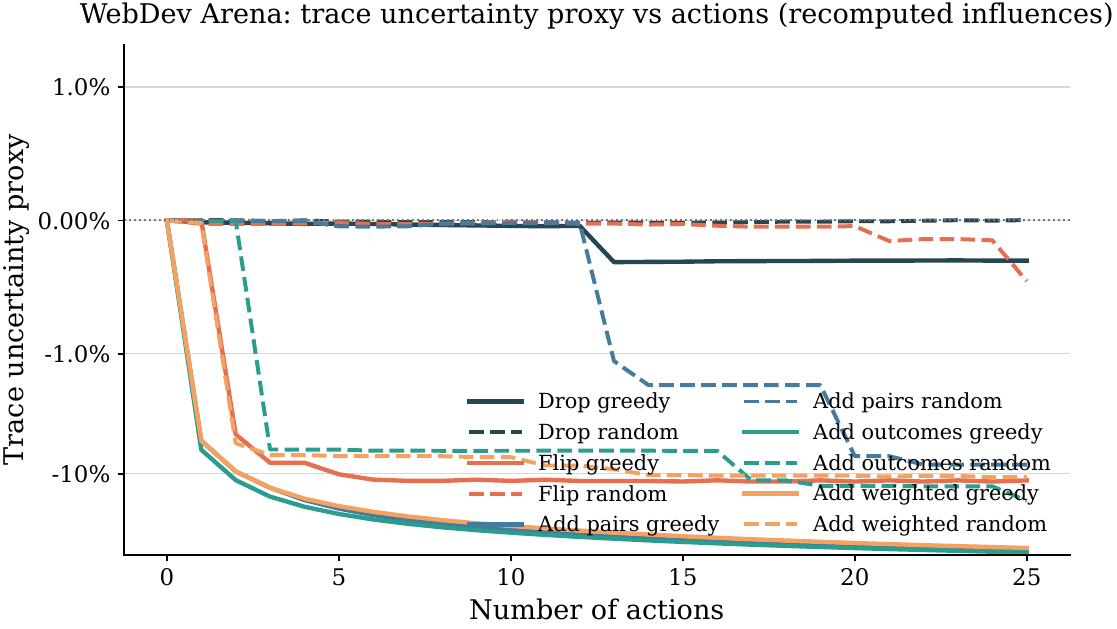}
        \caption{Trace uncertainty (\%).}
    \end{subfigure}
    \caption{Perturbation-budget curves for \textbf{WebDev Arena} (10{,}501 matches). The sparse graph results in severe $\tau$ degradation ($\approx 0.65$ under Flip) while Add variants still dominate uncertainty.}
    \label{fig:app_webdev}
\end{figure}

Figures~\ref{fig:app_llm_judge}--\ref{fig:app_webdev} report the full perturbation-budget curves for other datasets beyond Arena 55k (shown in the main paper, Figure~\ref{fig:global_ci_curves}). The same qualitative patterns hold throughout: influence-guided \Flip causes the steepest $\tau$ degradation, \Add greedy variants dominate trace uncertainty reduction, and random selection is negligible in every case. Datasets with sparser comparison graphs (ATP) exhibit much larger absolute $\tau$ degradation---reaching near 0 in the ATP case---confirming that graph density is a key determinant of global robustness.

\subsection{Top-\texorpdfstring{$k$}{k} Membership Change: CI-Aware vs. Non-CI-Aware}
\label{sec:topk-ci-vs-nonci}

\begin{figure}[t]
    \centering
    \begin{subfigure}{0.48\linewidth}
        \centering
        \includegraphics[width=\linewidth,trim=0 0 0 9mm, clip]{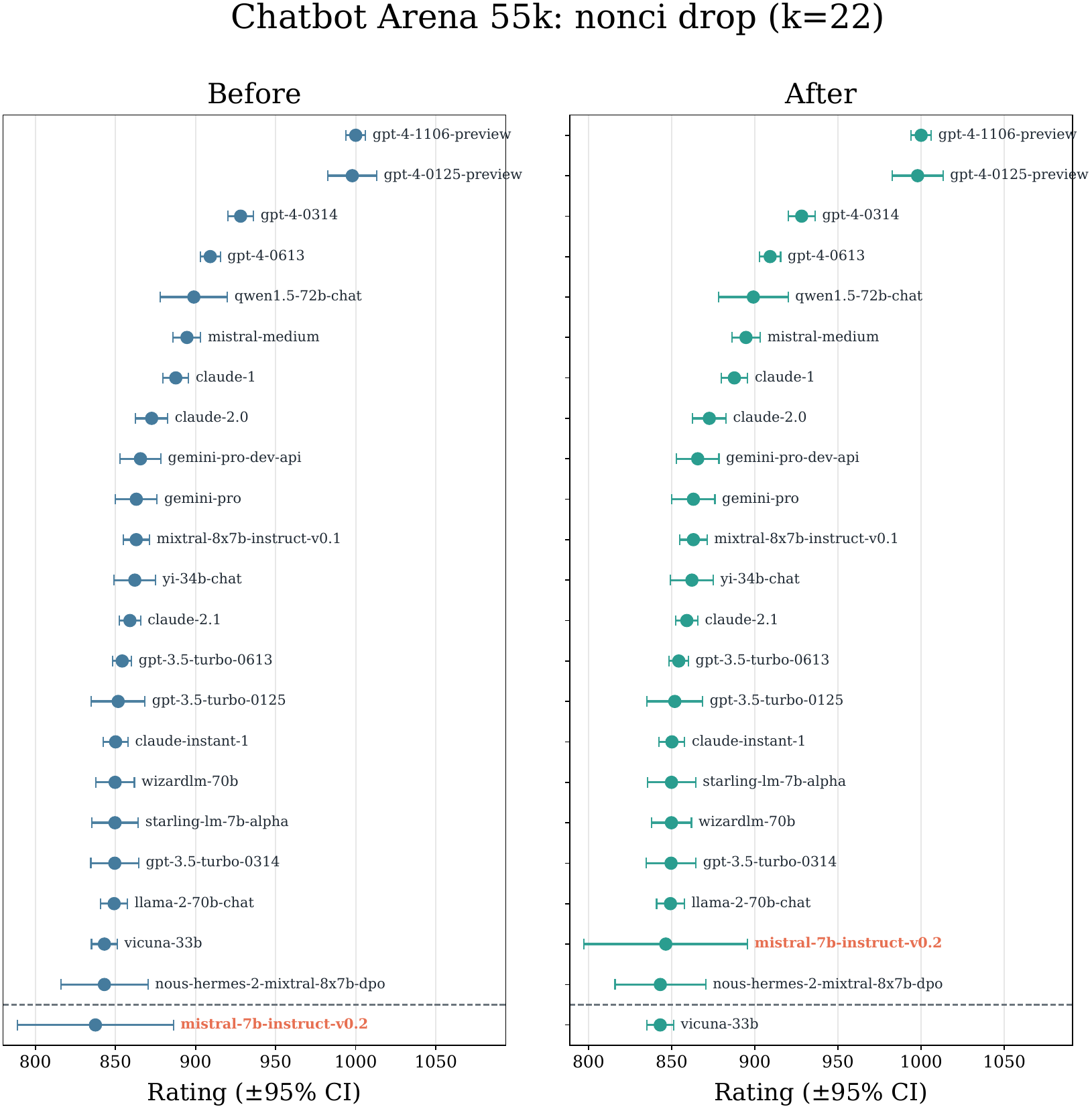}
        \caption{Non-CI-aware Top-$k$ change.}
        \label{fig:arena55k-k22-nonci-drop}
    \end{subfigure}
    \hfill
    \begin{subfigure}{0.48\linewidth}
        \centering
        \includegraphics[width=\linewidth,trim=0 0 0 9mm, clip]{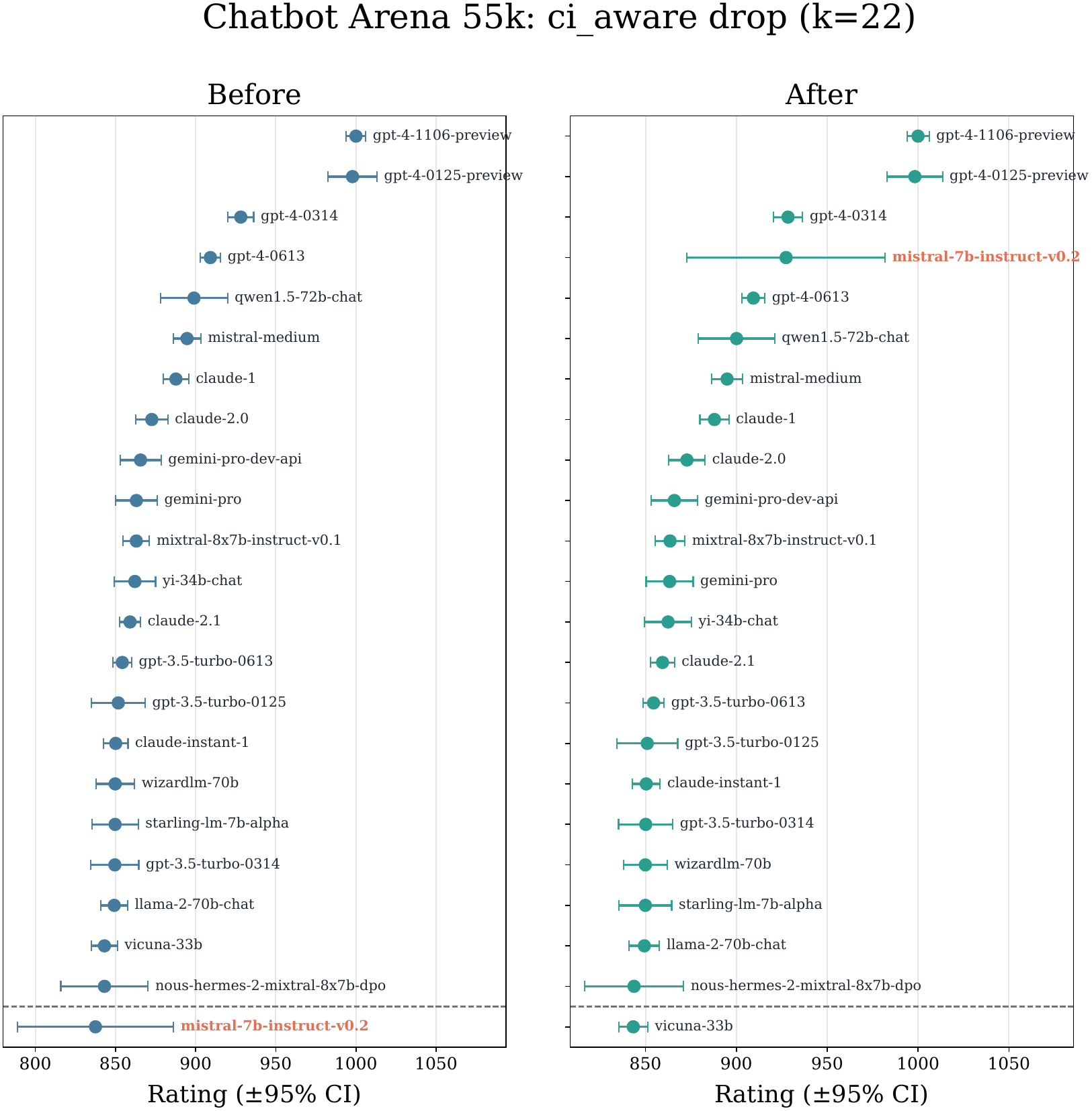}
        \caption{CI-aware Top-$k$ change.}
        \label{fig:arena55k-k22-ci-drop}
    \end{subfigure}
    \caption{
Top-$k$ membership change under point-estimate and CI-aware objectives on Arena 55k.
For $k=22$, influence-guided \Drop{} changes the point-estimate Top-$k$ boundary after
2 actions, while the stricter CI-aware criterion requires 19 actions for the outsider
to overtake the insider with non-overlapping confidence intervals. The two panels show
that boundary membership is fragile, yet CI-aware success requires a more statistically
robust change.
}
    \label{fig:arena55k-k22-ci-vs-nonci-drop}
\end{figure}

Figure~\ref{fig:arena55k-k22-ci-vs-nonci-drop} compares the effect of
influence-guided drop perturbations on Top-$k$ membership under the standard
point-estimate objective and the stricter CI-aware objective. In the
non-CI-aware setting, a membership change is counted as soon as the estimated
skill ordering crosses the Top-$k$ boundary, even if the affected models have
overlapping confidence intervals; on Arena 55k with $k=22$, this occurs after
only 2 targeted \Drop{} actions. By contrast, the CI-aware setting requires a
stronger form of manipulation: the promoted model must not only cross the
boundary in point estimate, but must do so with sufficient statistical separation
from the displaced model, requiring 19 targeted \Drop{} actions in the same
setting. The comparison therefore separates two notions of robustness:
instability of the displayed ranking versus instability that remains significant
after accounting for uncertainty. As expected, CI-aware Top-$k$ changes require
stronger perturbations, but the fact that targeted drops can still alter
membership under this stricter criterion shows that the leaderboard is not only
locally sensitive but can also be vulnerable in a statistically meaningful sense.

\section{Ablation}
\label{app:other-experiments}

\subsection{Specification of matches}
\label{app:match-specification}
\begin{figure}[H]
      \centering
      \includegraphics[width=\linewidth,trim=0 0 0 6mm, clip]{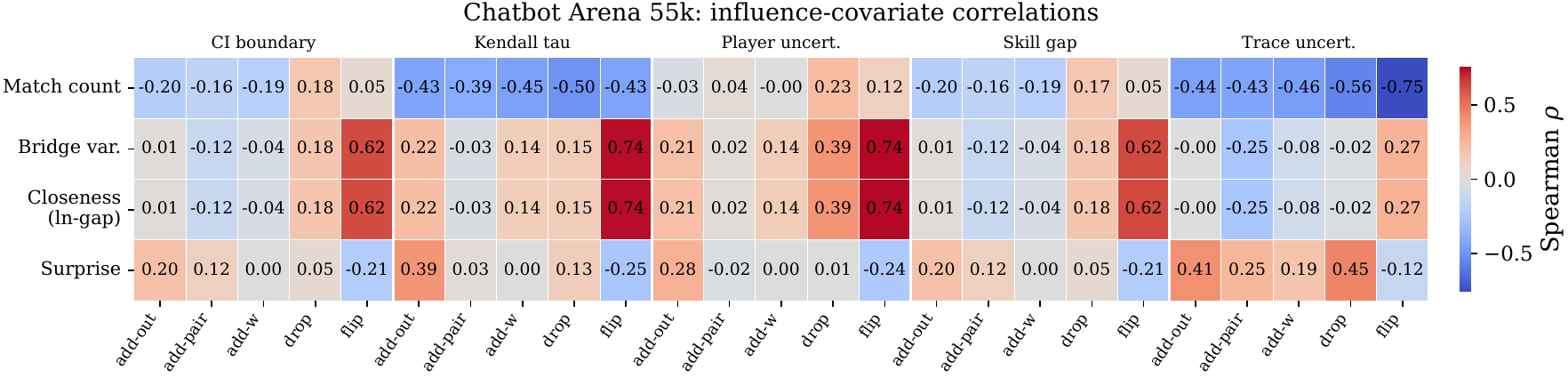}
      \caption{Signed Spearman correlations between one-step match influence scores and four match-level covariates
  in Chatbot Arena 55k, grouped by objective family and action type. Across most objectives, match count is weakly
  to strongly negative, while bridge variance and closeness are especially large for flip, highlighting the
  importance of uncertain, tightly contested comparisons in determining influential matches.}
      \label{fig:arena55k-match-spec-corr}
  \end{figure}

\paragraph{Influential matches reflect structure, not exposure.}
Figure~\ref{fig:arena55k-match-spec-corr} reports correlations between one-step match effects and
simple match-level covariates across match-specification objectives. We consider four covariates:
\textit{match count}, the number of times the same unordered pair appears in the training data;
\textit{bridge variance}, a match-level structural diversity score computed from the variance of the
opponents faced by the two endpoint players, intended to capture whether the match connects players
with broad or heterogeneous comparison neighborhoods; \textit{closeness}, the log-transformed
absolute BT skill gap between the two matched players; and \textit{surprise}, the discrepancy between
the observed outcome and the probability predicted by the fitted BT model. Because the natural
direction of degradation differs across objectives, decreasing \KTau{} indicates worse global ranking
stability, whereas increasing trace uncertainty indicates worse uncertainty, we interpret signed
correlations relative to each objective rather than as a universal measure of influence magnitude.
Under this objective-specific interpretation, \textit{match count} is not a stable proxy for influence:
frequently observed pairs are not consistently the ones whose perturbation most harms the leaderboard
objective. By contrast, \textit{bridge variance} and \textit{closeness} often become strongly associated
with large effects under flip objectives, suggesting that influence concentrates on structurally important
and closely contested comparisons, where reversing an outcome can propagate broadly through the
ranking. The \textit{surprise} covariate is more heterogeneous, with its association changing across add,
drop, and flip actions, especially for \KTau{}, player uncertainty, and trace uncertainty. Overall, the
figure suggests that influential matches are objective-dependent and are shaped more by structural
uncertainty and near-boundary comparisons than by raw exposure alone.

\paragraph{Influence concentrates on specified competitors.}
Under match-specific objectives, influence rankings concentrate sharply on the intended entities. Every objective that explicitly targets a player or player pair achieves a focus-hit rate of $1.0$ at top-20 selected action across all five action types: \Drop, \Flip, \Add-Pairs, \Add-Outcomes, and \Add-Weighted. We test this on Arena55k, for the CI-boundary and skill-gap objectives defined on \texttt{gpt-4-1106-preview} versus \texttt{gpt-4-0125-preview}, as well as the player-uncertainty objective for \texttt{gpt-4-1106-preview}, all top-20 influential matches involve the designated focal player(s). Thus, when the objective specifies which competitors or matchups matter, influence does not spread diffusely across the dataset; it identifies edits tightly aligned with the target player or pair. In contrast, global objectives such as \KTau and trace uncertainty have no designated focal player, so this focus-hit metric is not applicable to them.


\subsection{Effect of \texorpdfstring{$k$}{k} on \TopkMem}
\label{app:budget-k}
\begin{figure}[H]
    \centering
    \includegraphics[width=\linewidth,trim=0 0 0 6mm, clip]{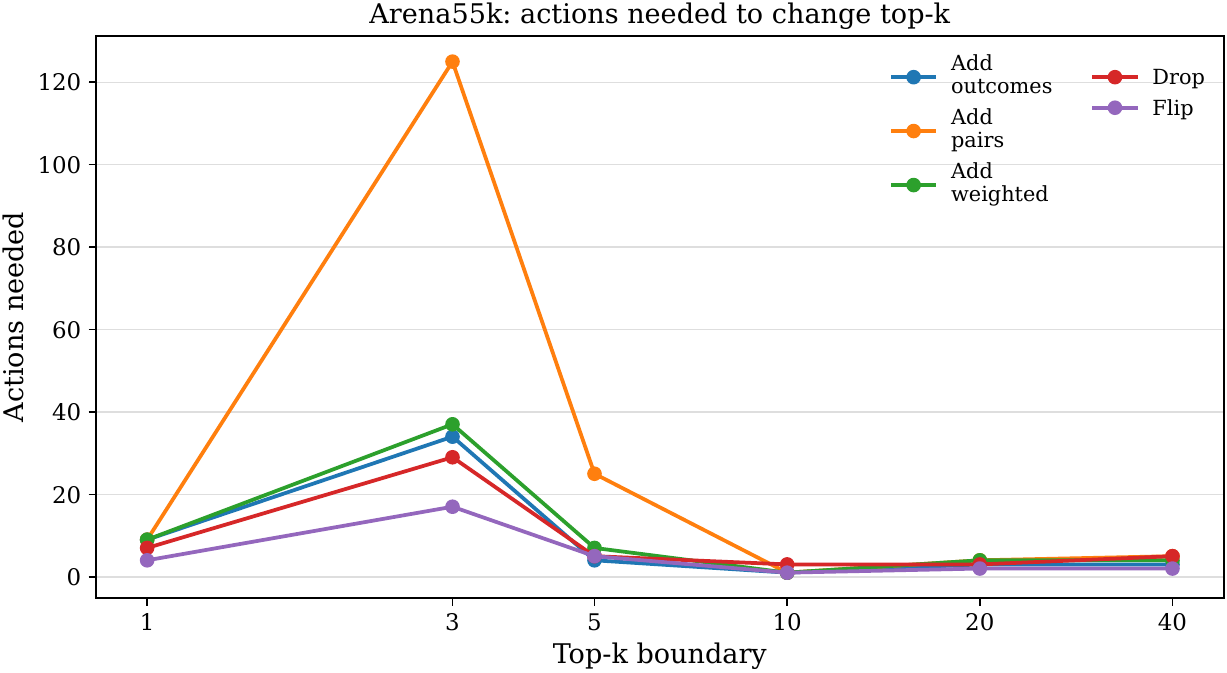}
\caption{
{Top-$k$ robustness is boundary-dependent.}
Minimum number of influence-guided actions required to change the Top-$k$ set
on Chatbot Arena 55k as $k$ varies.
The Top-3 boundary is the most structurally resistant, requiring roughly 120
actions, while several larger-$k$ boundaries require only a few actions.
Thus, robustness is not monotonic in $k$ but depends on the local structure of
the leaderboard near each cutoff.
}
    \label{fig:arena55k-topk-actions}
\end{figure}

To study how the choice of k affects the amount of data needed to change \TopkMem{}, we vary k and measure the required perturbation budget. Figure~\ref{fig:arena55k-topk-actions} shows that the effort required to change the top-$k$ ranking in Arena55k
  is highly non-monotonic in $k$ and depends strongly on the action type. The most striking feature is the sharp
  peak at $k=3$, where adding pairs becomes dramatically more expensive than all other interventions, while adding
  weighted outcomes, adding outcomes, and dropping results remain substantially smaller, and flipping outcomes is
  consistently the least costly. Beyond $k=5$, the required number of actions drops quickly for all methods and
  remains low and fairly stable through $k=10,20,40$, indicating that deeper top-$k$ boundaries are easier to
  perturb than the top few ranks. Overall, the figure suggests that the top-3 boundary is the most structurally
  resistant part of the ranking, especially when interventions are constrained to adding new comparisons rather
  than altering existing ones.

\subsection{Structural predictors of player-removal impact}
\label{app:player-feature-association}

\begin{figure}[H]
    \centering
    \includegraphics[width=0.62\linewidth,trim=0 0 0 7mm, clip]{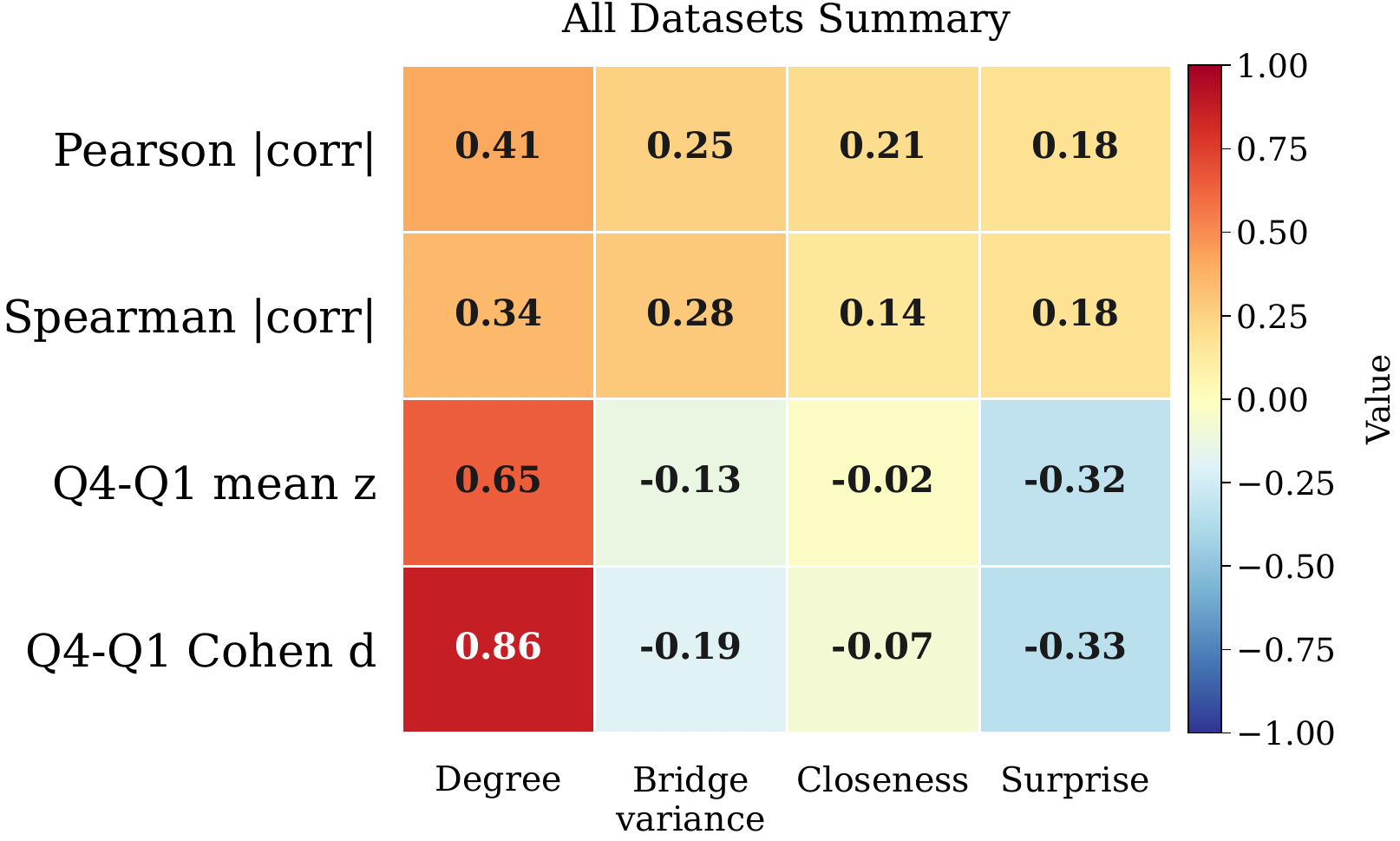}
    \caption{%
    Association between structural player features and absolute $|\tau|$-influence, 
    aggregated across datasets using Pearson and Spearman correlations, Q4--Q1 mean 
    $z$-score, and Cohen's $d$. Degree is the strongest and most consistent predictor 
    of player-removal impact.
    }
    \label{fig:kendall_feature_association}
\end{figure}

To understand what drives player-removal impact, 
Figure~\ref{fig:kendall_feature_association} reports four association statistics 
between each structural player feature and absolute $|\tau|$-influence, 
aggregated across datasets. We consider {\textit{degree}}, the number of matches a 
player participates in; \textit{bridge variance}, how unevenly a player connects 
otherwise weakly connected regions of the graph; \textit{closeness}, how 
centrally located a player is in terms of shortest-path distances; and 
\textit{surprise}, how unexpected a player's outcomes are relative to model 
predictions.

Degree is the strongest and most consistent predictor: it achieves the highest 
Pearson ($|r|=0.41$) and Spearman ($|\rho|=0.34$) correlations, and the largest 
quartile separation (Q4--Q1 mean $z=0.65$, Cohen's $d=0.86$). Bridge variance 
and closeness show weaker or less consistent associations, while surprise is 
only weakly associated.

\newpage

\end{document}